\documentclass[fleqn,10pt]{wlscirep}
\usepackage[utf8]{inputenc}
\usepackage[T1]{fontenc}
\usepackage{lineno}
\usepackage{float}
\usepackage{hyperref}

\usepackage{csquotes}
\usepackage{float}
\usepackage{svg}
\usepackage{multirow}  
\usepackage{tikz}
\usepackage{graphicx}
\usepackage{subcaption}
\usetikzlibrary{arrows.meta, positioning, fit, backgrounds, calc}
\usepackage[T1]{fontenc}
\usepackage{helvet}
\usepackage{courier} 
\usepackage{forest}

\usepackage{listings}
\usepackage{xcolor}

\lstdefinestyle{jsonstyle}{
  basicstyle=\ttfamily\small,
  breaklines=true,
  frame=single,
  showstringspaces=false,
  stringstyle=\color{blue!60!black},
  commentstyle=\color{gray},
  keywordstyle=\color{black},
}

\usepackage{colortbl}
\usepackage{xcolor}

\usepackage{longtable}
\usepackage{colortbl}
\usepackage{xcolor}
\usepackage{booktabs}

\restylefloat{table}

\title{Mapping how LLMs debate societal issues when shadowing human personality traits, sociodemographics and social media behavior}

\author[1,*]{Ali Aghazadeh Ardebili}
\author[1]{Massimo Stella}
\affil[1]{CogNosco Lab, University of Trento, Department of Psychology and Cognitive Science, Trento, Italy}
\affil[*]{Corresponding author: a.a.ardebili@unitn.it}

\begin{abstract}

Large Language Models (LLMs) can strongly shape social discourse, yet datasets investigating how LLM outputs vary across controlled social and contextual prompting remain sparse. Cognitive Digital Shadows (CDS) is a 190,000-record synthetic corpus supporting analyses of LLM-generated discourse. Each CDS record is generated by one of 19 LLMs, prompted to shadow either a human persona or an AI-assistant role. CDS contains LLM responses on 4 controversial societal topics: vaccines/healthcare, social media disinformation, the gender gap in science, and STEM stereotypes. Persona-conditioned records encode 17 sociodemographic and psychological attributes, providing data linking LLMs' prompts, language, stances and reasoning. Texts are validated for topic anchoring and can support emotional analyses via interpretable NLP (e.g. textual forma mentis networks). CDS is enriched by a pooling platform with user-friendly dashboards, enabling easy, interactive group-level comparisons of emotional and semantic framing across personas, topics and models. The CDS prompting framework supports future audits of LLMs' bias, social sensitivity and alignment.

\end{abstract}
\begin{document}

\flushbottom
\maketitle

\thispagestyle{empty}

\section*{Background \& Summary}


Large language models (LLMs) are rapidly becoming integral components of public information environments \cite{rossetti2024social,branda2026comfort,DeDuro2025}. These LLMs provide summaries of news articles \cite{zhang2024benchmarking}, answers to health questions \cite{branda2026comfort}, content moderation \cite{ferrara2026generative} and also join discussions on divisive topics in society \cite{DeDuro2025}. As language models transition from mere knowledge retrieval devices \cite{bender2021dangers} into proactive agents \cite{Chen2026,chen2026openclaw} (OpenClaw bots, for example), an important research question arises:

\begin{center}
\textit{What prompt- and model-dependent patterns shape LLM-generated argumentation and positioning on socially sensitive topics?}
\end{center}

Benchmarks currently measure either accuracy, toxicity or how well models follow instructions \cite{bender2021dangers,weidinger2022taxonomy}, yet they overlook socio-cognitive dynamics in opinion formation, negotiation and expression within LLMs \cite{DeDuro2025}. Psychological research spanning several decades demonstrates that human discourse is driven by personal factors, such as individual identity, personality and attention \cite{pennycook2021psychology,shapiro2012role}. For instance, opinions about vaccines, misinformation, or gender equality do not arise in a vacuum but rather reflect psychological dispositions, demographic position, social exposure, and media consumption patterns \cite{pennycook2020fighting,guess2019less}. 

\textbf{Gaps in LLMs as complex systems.} This multidimensional landscape calls for a complex systems approach \cite{ardebili2021digitaltwins,ardebili2023navigating,abramski2023cognitive}, studying LLMs' behavior  in socially realistic conditions, either aligned with human-inspired features or relative to the GenAI nature of LLMs \cite{stella2023using}. This complexity calls for representations that preserve the actual properties of the system and its cognitive and communicative patterns without oversimplifying them, while remaining open to rigorous analysis. Such representations enable data-driven methods to systematically capture the system's behavior  across conditions. This is a requirement that aligns closely with the defining characteristics of a \textit{Digital twin (DT)}, making it a natural candidate solution. A DT is a dynamic virtual replica of a real-world system that maintains continuous synchronisation with its physical counterpart, enabling bidirectional interaction and real-time state correspondence \cite{Jones2020,Ardebili2021EIOT,Barricelli2019,aghazadeh2024digital}. However, while a DT offers precisely this kind of representational power, a \textit{digital shadow} provides a more parsimonious alternative when only a one-way observational log of a system's behavior  across conditions is needed — capturing what the system does without modeling or influencing it \cite{GAFFINET2025104230,Singh2021,ardebili2021digitaltwins,Bergs2021}. In sum, rather than simulating the system, a digital shadow captures its traces, making it especially suitable for representing the cognitive and communicative patterns of LLMs \cite{Voronin2025, DeDuro2025, abramski2023cognitive} across socially realistic scenarios. However, digital-shadow concepts remain underexplored in cognitive and language-based systems, representing a significant gap in the literature that this work aims to address.

To close this gap in research, we present the \textit{Cognitive Digital Shadows} (CDS) dataset by incorporating the technical notion of digital shadows \cite{sepasgozar2021differentiating, aghazadeh2024digital} through a cognitive/psychological lens \cite{abramski2023cognitive,stella2023using}. Specifically, we present CDS as a comprehensive synthetic dataset that examines how output produced by LLMs changes in response to structured persona conditioning and AI-self role-playing for socially sensitive topics. In doing so, CDS is meant to be used as a tool for analyzing prompt-based discourse, bias sensitivity, and representations of text generation in general. At the same time, CDS is not intended to represent actual human attitudes or beliefs held within society. The cognitive shadow is purely observational in nature \cite{ardebili2023navigating}: It preserves the generated output and its metadata without ever updating or affecting any other system or asserting the existence of a bidirectional link between the two systems. This definition distinguishes CDS from digital twin frameworks and aligns the dataset with one-way observation of model behavior across controlled conditions. 

\textbf{Operational definition of a Cognitive Digital Shadows (CDS).} We define a CDS as the structured trace of an LLM-generated response produced under a specified contextual configuration. In human-simulated mode, this configuration includes a synthetic persona described by psychologically and socially meaningful attributes. In AI-assistant mode, it includes only the topic and role instruction. The shadow is therefore the generated response plus its metadata. When human-simulated, the shadow includes psychological traits (e.g., OCEAN personality profiles \cite{john1999big}, anxiety \cite{lovibond1995structure}, curiosity \cite{lydon2021hunters}), biological characteristics (e.g., age, sex, etc. \cite{john1999big}), and sociodemographic features (e.g., occupation, education, income proxies, location, hobbies, and social media exposure \cite{zou2018ai}). CDS includes these features because of past results across social cognition research and computational social science, showing how all these human features can influence arguments, perceptions and personal opinions. More in detail, personality traits are known to influence communication style and conflict behavior  \cite{john1999big}. Sociodemographic positioning shapes information exposure and trust in institutions \cite{guess2019less}. Attention environments, including time spent on social media, can affect susceptibility to misinformation \cite{pennycook2020fighting,pennycook2021psychology}. By embedding these constructs directly into LLMs' prompts, the Cognitive digital shadows dataset operationalizes a controlled experimental environment for studying context-conditioned language generation in large language models. In other words, the CDS prompting framework explicitly includes human-inspired attributes in the prompt, allowing systematic observation of how these contextual variables are associated with variation in LLM-generated argumentative text. Cognitive Digital Shadows can represent individual human features but, as digital shadows \cite{ardebili2021digitaltwins}, they are not intended to influence real people. This lack of influence is due to the fact that Cognitive Digital Shadows are simulated in isolation from real people, unlike other platforms where multiple agents can be simulated and interact with each other, like on YSocial \cite{rossetti2024social}, or with other humans, like OpenClaw bots \cite{chen2026openclaw}. Cognitive Digital Shadows are rather theoretical instruments that make latent social and psychological variables explicit and manipulable when inspired by human-based research in psychology and cognitive science. This transparency enables controlled descriptive comparisons across prompt configurations, supporting reproducible audits of how model outputs vary when contextual attributes are changed \cite{liu2023trustworthy}. These comparisons should not be interpreted as causal effects unless the relevant attributes are experimentally isolated and analyzed with an appropriate design. The framework also supports the study of emergent biases, such as identity-based stereotyping \cite{abramski2023cognitive,shapiro2012role} or inconsistent public-health framing across demographic contexts \cite{branda2026comfort,pennycook2021psychology}.

\textbf{Dataset scope and motivation.} Our CDS dataset covers four high-salience societal domains: (i) management of fake content on social media \cite{briand2021infodemics}, (ii) the gender gap in science \cite{stella2020text}, (iii) stereotype threat in STEM \cite{shapiro2012role}, and (iv) public discussions about healthcare and COVID-19 vaccines \cite{pennycook2020fighting}. These topics were selected because they combine empirical uncertainty, moral reasoning, and identity relevance \cite{pennycook2021psychology,stella2020text}. Furthermore, these topics are frequently encountered in online discourse environments that can be highly polluted by social bots or LLM-powered fake users \cite{ferrara2026generative}, contributing to epistemic ecosystems where AIs might be influencing humans \cite{Branda2026-BRAWAI-5,coppolillo2026mosaic,briand2021infodemics,rossetti2024social}. Each data instance contains structured metadata and a generated position expressed as continuous prose, together with reasoning summaries and contextual descriptors (e.g., tone, referenced persona fields). The dataset's dual design enables comparisons across two modes of social cognition: (i) simulated human personas with heterogeneous demographic and psychological configurations, and (ii) AI-reflexive agents that reason as large language models about evidence, uncertainty, and institutional knowledge. The combination of human-like and AI-like personas can capture an increasingly relevant interaction paradigm where LLMs present themselves as neutral or data-driven assistants \cite{rossetti2024social}.

\textbf{User-friendly features of CDS.} One major aspect of CDS is the presence of a pooling and aggregation module, tailored for use by psychologists, social scientists, and political scientists with little experience in creating and processing big datasets. With the totally automated pooling facility, instances of personas can be searched based on the distribution of topics and attributes. It would allow researchers to formulate statistical perception profiles of pools or groups of personas within an LLM – e.g., \textit{"How did all DeepSeek 3.2's personas who lived in New York and were employed as teachers talk about the gender gap?"} Using the pooling system in CDS, scientists can rapidly estimate general trends \cite{DeDuro2025} such as changes in tone, epistemic certainty, or value-ladenness in all demographic categories, personalities, or exposure to media. The system helps conduct group-level descriptive analyses over synthetic prompt configurations \cite{de2025measuring}, but still allows for complete traceability of each instance individually. This method facilitates the exploration of various research questions relevant to scientists studying the behavior and psychology of LLMs \cite{binz2023using,coppolillo2026mosaic,rossetti2024social}, such as, for example:
\begin{enumerate}
    \item whether high-neuroticism profiles elicit more cautious language in LLMs as in humans \cite{mehl2013taking};
    \item whether high social-media exposure increases polarization framing in LLMs as in humans \cite{rossetti2024social,ferrara2026generative};
    \item AI-self responses systematically emphasize institutional authority like humans trusting institutions can do \cite{pennycook2021psychology,stella2020text}.
\end{enumerate}

\textbf{Scope of the dataset.} Cognitive Digital Shadows dataset offers an easy-to-reuse dataset for studying how LLM-generated responses vary across psychologically and socially informed prompt conditions. The CDS dataset is rooted in cognitive science \cite{binz2023using,Branda2026-BRAWAI-5,stella2023using}, social psychology \cite{john1999big,lovibond1995structure,pennycook2021psychology}, digital shadowing, and computer science \cite{zou2018ai,chen2026openclaw,ardebili2023navigating}. Hence, the dataset is intended to enable future reproducible studies of model sensitivity, framing, and bias in socially sensitive prompt contexts. In addition, it is vital to highlight that CDS is a dataset for evaluating the language generated by LLMs based on prompt conditions, and it is not a surrogate for human opinions. It was developed to examine the impact of structured persona descriptions and AI self-roles on the generated content when exposed to certain prompt conditions. Thus, the dataset facilitates comparisons between the sensitivity of different models to various contextual features. However, the dataset does not claim to generate accurate representations of real people or demographics.

\section*{Methods}
\begin{figure}[h!]
    \centering
    \includegraphics[width=0.88\linewidth]{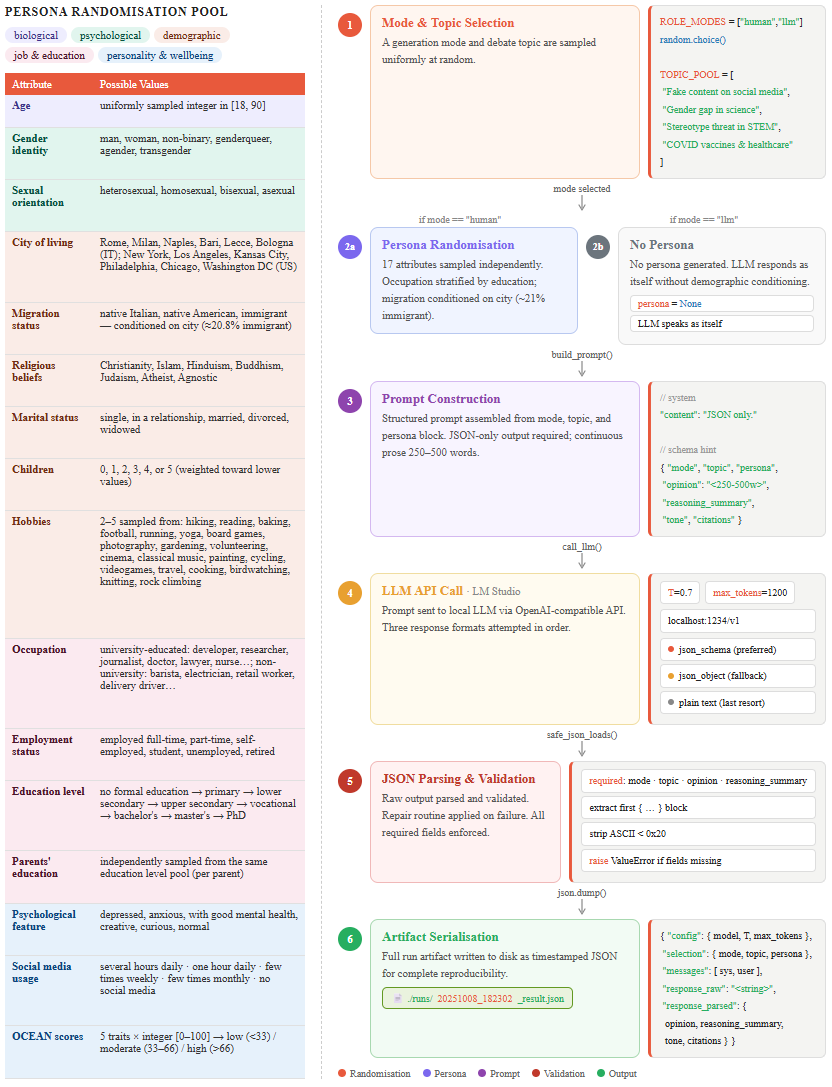}
    \caption{Cognitive Digital Shadow Generation Pipeline (\texttt{run\_once()}). Step~1 randomly selects a generation mode and debate topic. Step~2 either constructs a fully specified synthetic sociodemographic persona (human mode) or assigns none (LLM mode). Step~3 assembles a structured prompt requiring strict JSON-only output. Step~4 submits the prompt to a OpenAI-compatible inference endpoint, local or provider-hosted via an OpenAI-compatible API. Step~5 parses and validates the raw output, enforcing completeness of all required fields. Step~6 serializes the full artifact as a timestamped JSON file.}
    \label{fig:pipeline_generation}
\end{figure}

This section describes the full pipeline used to generate, process, and analyze the 190,000 entries in the Cognitive Digital Shadows corpus. The pipeline for producing the CDS dataset is illustrated in Figure \ref{fig:pipeline_generation}. We outline the LLM-based data generation procedure across the following steps. We start by discussing our psycho-social personification scheme (Figure \ref{fig:pipeline_generation} left, "Persona Randomisation Pool"). Then we describe the pre-processing steps applied to the raw text (Figure \ref{fig:pipeline_generation} right, steps 1-6). Finally, we outline the validation process through interpretable natural language processing, i.e. textual forma mentis networks \cite{stella2020text,haim2026cognitive,Semeraro2025}.

\subsection*{Data generation}\label{sub1}
The pipeline was executed across 19 different LLMs, including both locally hosted and API-based models (Claude Sonnet 4.5 and DeepSeek V3.2), each generating a minimum of 10{,}000 validated records, yielding a large-scale corpus of synthetic debates covering 4 controversial topics. Records were discarded during the data cleaning stage if they failed structural completeness checks, e.g. incomplete JSON files or missing fields. Completeness was assessed also on human-simulated responses (e.g. incomplete persona attributes) and applied to both human and LLM-generated data for topic coverage, opinion, and reasoning summary corpus. The resulting discard rate varied across models, ranging from 0\% ($DeepSeek-V3.2_Model$ and $llama3.3-70B$) to a maximum of 15.3\% ($LiquidAI-LFM2-1.2B$). Fig. \ref{fig:pipeline_generation} illustrates this pipeline. Every generated structured document (a JSON file) was validated for structural completeness: all records, regardless of generation mode, were required to contain a topic identifier, a continuous-prose opinion, and a reasoning summary. 

Records generated in human-simulated mode were additionally required to carry a fully populated sociodemographic persona, with no missing attributes across 17 fields spanning demographic, geographic, socioeconomic, family, psychological, and 
personality dimensions (see Section \textbf{Data Records}). Any structured document failing these completeness checks was discarded and excluded from the final corpus, ensuring that downstream analyses are conducted on a clean, structurally consistent dataset.

\subsubsection*{Prompt building: Persona Randomisation Pool}
Prompt construction followed established best practices in prompt engineering \cite{jacobsen2025promises,binz2023using} to ensure methodological rigor and reproducibility. Output format was enforced by embedding a concrete JSON schema as an in-context exemplar, a technique shown to substantially improve format compliance in AI and ML outputs \cite{vertexai_multimodal_control}. 

Importantly, when \texttt{mode = "human"}, the prompt injects a richly specified persona comprising age, biological sex, gender identity, sexual orientation, occupation, city of residence, education level, parental education, marital status, migration status, psychological features, religious beliefs, social media usage \cite{kross2021social}, hobbies, and OCEAN personality trait scores \cite{Mahale2025,john1999big}. The use of OCEAN (Big Five) personality scores as prompt conditioning variables follows established practices in the literature, including personality prediction through the OCEAN model (comprising Openness, Conscientiousness, Extraversion, Agreeableness, and Neuroticism) \cite{Mahale2025}, the use of Big Five traits in generating AI-driven empathetic responses \cite{Lalwani2025}, and the emulation of personality traits in multi-agent systems \cite{Liapis2024}. 

Building on these past empirical results of a correspondence between LLMs' behavior  and injected personality traits, we expected the insertion of personality traits in the personification prompt to increase controlled variation in generated responses and condition its output variance, leading to better Cognitive Digital Shadows (i.e. more varied responses).

In providing textual responses about societal topics, CDS' prompt specifies a word range of 250--500 words and explicitly prohibits bullet points, noting that explicit length and format constraints reduce variance in model outputs and improve suitability for structured downstream analysis. 

Before scaling up the data simulation process, robustness to temperature tuning was tested on the DeepSeek-r1 70b \cite{guo2025deepseek}, which represents a medium-sized language model demonstrating adequate accuracy in multiple decision-making and NLP applications. Tuning of $T$ was carried out in steps of $\delta = 0.1$ from 0.0 to 1.0 with a particular AI assistant personification. For each $T$, 100 experiments were run in order to measure output length, JSON validity and lexical diversity of opinion fields. Given that $T = 0.7$ offered a good trade-off between these aspects, we chose it as the target temperature setting.

Analogous results were obtained when the temperature was tuned over an equivalent range but for a randomly selected human personification. On the basis of this robustness check, we considered temperature did not produce obvious failures in this pilot the responses provided by the model and hence we fixed all simulations for CDS at an intermediate value $T=0.7$, high enough to guarantee a varied search of lexical/semantic output space while reducing the occurrence of hallucinations or incoherent text generation \cite{peeperkorn2024temperature}. 

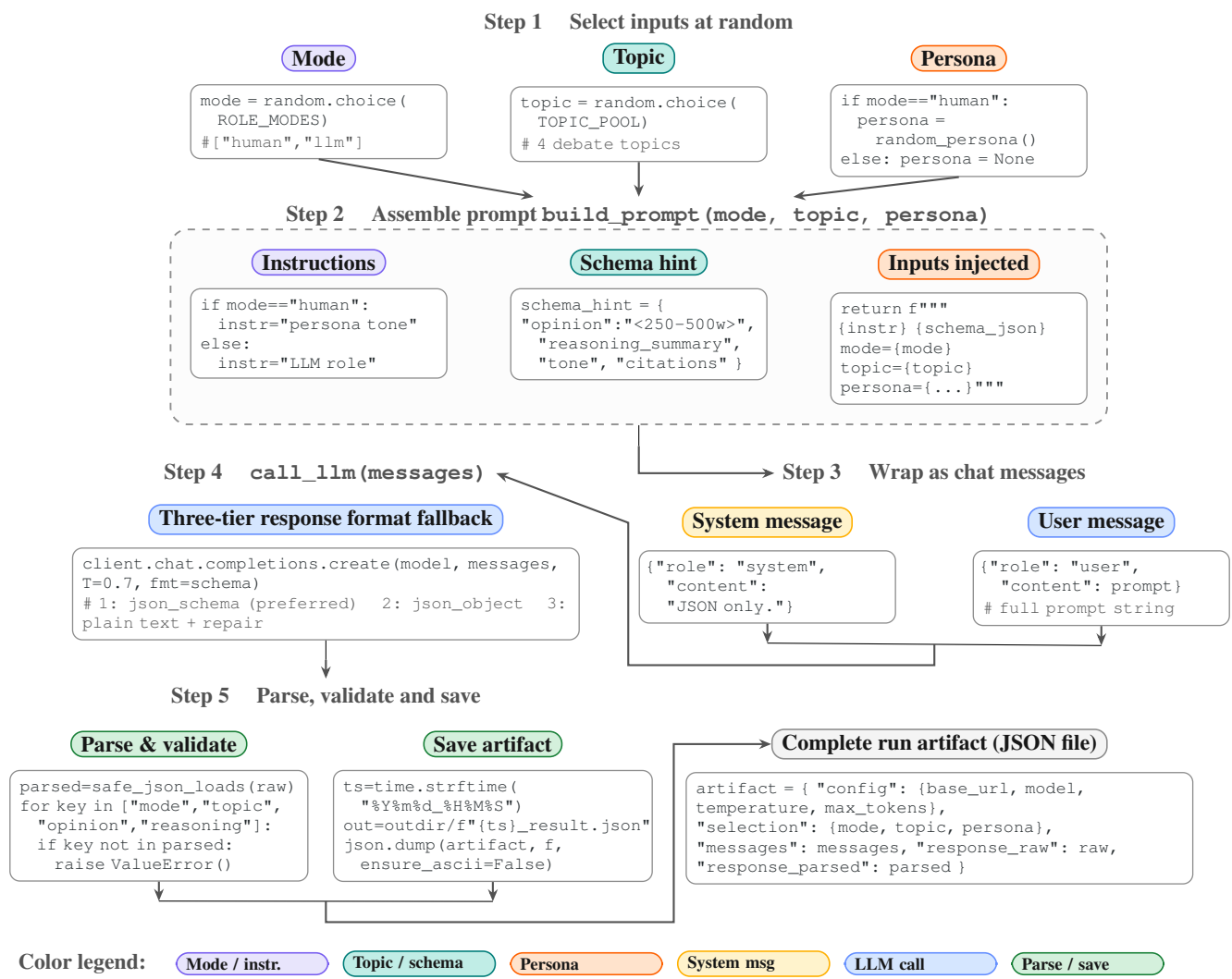
\begin{figure}[h!]
    \centering

\definecolor{ibmVioletFill}  {HTML}{E8E5FC}
\definecolor{ibmVioletBorder}{HTML}{785EF0}
\definecolor{ibmTealFill}    {HTML}{C0EDE6}
\definecolor{ibmTealBorder}  {HTML}{007D79}
\definecolor{ibmOrangeFill}  {HTML}{FFE3CC}
\definecolor{ibmOrangeBorder}{HTML}{FE6100}
\definecolor{ibmYellowFill}  {HTML}{FFF3CC}
\definecolor{ibmYellowBorder}{HTML}{FFB000}
\definecolor{ibmBlueFill}    {HTML}{D5E6FF}
\definecolor{ibmBlueBorder}  {HTML}{648FFF}
\definecolor{ibmGreenFill}   {HTML}{D4F0DC}
\definecolor{ibmGreenBorder} {HTML}{198038}
\definecolor{ibmGrayFill}    {HTML}{F4F4F4}
\definecolor{ibmGrayBorder}  {HTML}{8D8D8D}
\definecolor{textMain}       {HTML}{161616}
\definecolor{textSub}        {HTML}{525252}
\definecolor{codeColor}      {HTML}{21272A}
 
\tikzset{
  codebox/.style={
    rectangle, rounded corners=4pt,
    text width=3.4cm, align=left,
    font=\scriptsize\ttfamily, text=codeColor,
    inner sep=4pt, line width=0.5pt,
    fill=white, draw=ibmGrayBorder
  },
  codeboxWide/.style={
    rectangle, rounded corners=4pt,
    text width=9cm, align=left,
    font=\scriptsize\ttfamily, text=codeColor,
    inner sep=4pt, line width=0.5pt,
    fill=white, draw=ibmGrayBorder
  },
  colorbox/.style={
    rectangle, rounded corners=5pt,
    font=\small\bfseries, text=textMain,
    inner xsep=4pt, inner ysep=2pt, line width=0.6pt
  },
  sViolet/.style={colorbox, fill=ibmVioletFill, draw=ibmVioletBorder},
  sTeal/.style=  {colorbox, fill=ibmTealFill,   draw=ibmTealBorder},
  sOrange/.style={colorbox, fill=ibmOrangeFill, draw=ibmOrangeBorder},
  sYellow/.style={colorbox, fill=ibmYellowFill, draw=ibmYellowBorder},
  sBlue/.style=  {colorbox, fill=ibmBlueFill,   draw=ibmBlueBorder},
  sGreen/.style= {colorbox, fill=ibmGreenFill,  draw=ibmGreenBorder},
  sGray/.style=  {colorbox, fill=ibmGrayFill,   draw=ibmGrayBorder},
  sBluWide/.style= {colorbox, fill=ibmBlueFill, draw=ibmBlueBorder},
  sGrayWide/.style={colorbox, fill=ibmGrayFill, draw=ibmGrayBorder},
  container/.style={
    rectangle, rounded corners=7pt,
    draw=ibmGrayBorder, dashed, line width=0.7pt,
    fill=ibmGrayFill, fill opacity=0.25,
    inner sep=8pt
  },
  steplbl/.style={font=\small\bfseries, text=textSub},
  arr/.style={-{Stealth[length=5pt,width=4pt]},
              line width=0.8pt, color=textSub},
}

\begin{tikzpicture}[node distance=0.2cm and 0.3cm]
 
\node[steplbl,yshift=0.5cm] (s1h) {Step 1 \quad Select inputs at random};
 
\node[sViolet,  xshift=-4.6cm] (modeT) {Mode};
\node[codebox, below=0.18cm of modeT] (modeC)
  {mode = random.choice(\\
   \quad ROLE\_MODES)\\[1pt]
   \textcolor{textSub}{\#["human","llm"]}};
 
\node[sTeal] (topicT) {Topic};
\node[codebox, below=0.18cm of topicT] (topicC)
  {topic = random.choice(\\
   \quad TOPIC\_POOL)\\[1pt]
   \textcolor{textSub}{\# 4 debate topics}};
 
\node[sOrange, xshift=4.6cm] (personaT) {Persona};
\node[codebox, below=0.18cm of personaT] (personaC)
  {if mode=="human":\\
   \quad persona =\\
   \quad\quad random\_persona()\\
   else: persona = None};
 
\node[steplbl, below=0.5cm of topicC] (s2h)
  {Step 2 \quad Assemble prompt \texttt{build\_prompt(mode, topic, persona)}};
 
\node[sViolet, below=0.22cm of s2h, xshift=-4.6cm] (instrT) {Instructions};
\node[codebox, below=0.18cm of instrT] (instrC)
  {if mode=="human":\\
   \quad instr="persona tone"\\
   else:\\
   \quad instr="LLM role"};
 
\node[sTeal, below=0.22cm of s2h] (schemaT) {Schema hint};
\node[codebox, below=0.18cm of schemaT] (schemaC)
  {schema\_hint = \{
   \quad "opinion":"<250-500w>",\\
   \quad "reasoning\_summary",\\
   \quad "tone", "citations" \}};
 
\node[sOrange, below=0.22cm of s2h, xshift=4.6cm] (injT) {Inputs injected};
\node[codebox, below=0.18cm of injT] (injC)
  {return f"""\\
   \{instr\} \{schema\_json\}\\
   mode=\{mode\}\\
   topic=\{topic\}\\
   persona=\{...\}"""};
 
\begin{scope}[on background layer]
  \node[container,
        fit=(instrT)(instrC)(schemaT)(schemaC)(injT)(injC),
       ] (bpbox) {};
\end{scope}
 
\node[steplbl, right =0.1cm  of schemaC, yshift=-2cm] (s3h)
  {Step 3 \quad Wrap as chat messages};
 
\node[sYellow, below=0.22cm of s3h, xshift=-2.4cm] (sysT) {System message};
\node[codebox, below=0.18cm of sysT] (sysC)
  {\{"role": "system",\\
   \quad "content":\\
   \quad "JSON only."\}};
 
\node[sBlue, below=0.22cm of s3h, xshift=2.4cm] (userT) {User message};
\node[codebox, below=0.18cm of userT] (userC)
  {\{"role": "user",\\
   \quad "content": prompt\}\\[1pt]
   \textcolor{textSub}{\# full prompt string}};
 
\node[steplbl, left=4cm of s3h] (s4h)
  {Step 4 \quad \texttt{call\_llm(messages)}};
 
\node[sBluWide, below=0.18cm of s4h] (llmT)
  {Three-tier response format fallback};
\node[codeboxWide, below=0.18cm of llmT, text width=7cm] (llmC)
  {client.chat.completions.create(model, messages, T=0.7, fmt=schema)\\
   \textcolor{textSub}{\# 1: json\_schema (preferred) \quad 2: json\_object \quad 3: plain text + repair}};
 
\node[steplbl, below=0.55cm of llmC] (s5h)
  {Step 5 \quad Parse, validate and save};
 
\node[sGreen, below=0.22cm of s5h, xshift=-2.4cm] (parseT) {Parse \& validate};
\node[codebox, below=0.18cm of parseT, text width=4cm] (parseC)
  {parsed=safe\_json\_loads(raw)\\
   for key in ["mode","topic",\\
   \quad "opinion","reasoning"]:
   \\
   \quad if key not in parsed:\\
   \quad\quad raise ValueError()};
 
\node[sGreen, below=0.22cm of s5h, xshift=2.4cm] (saveT) {Save artifact};
\node[codebox, below=0.18cm of saveT, text width=4.3cm] (saveC)
  {ts=time.strftime(\\
   \quad"\%Y\%m\%d\_\%H\%M\%S")\\
   out=outdir/f"\{ts\}\_result.json"\\
   json.dump(artifact, f,\\
   \quad ensure\_ascii=False)};
 
\node[sGrayWide, right=3cm of saveT] (artT)
  {Complete run artifact (JSON file)};
\node[codeboxWide, below=0.18cm of artT , text width=7cm] (artC)
  {artifact = \{ "config": \{base\_url, model, \\temperature, max\_tokens\},\\
   "selection": \{mode, topic, persona\},\\
   "messages": messages, "response\_raw": raw, \\ "response\_parsed": parsed \}};
 
\draw[arr] (modeC.south)    -- (s2h);
\draw[arr] (topicC.south)   -- (s2h);
\draw[arr] (personaC.south) -- (s2h);
 
\coordinate (bpmid) at ($(bpbox.south)+(0,-0.3)$);
\draw (bpbox.south) -- (bpmid);
\draw[arr] (bpmid) -- (bpmid |- s3h.west) -- (s3h.west);
 
\coordinate (sysBot)  at ($(sysC.south)+(0,-0.28)$);
\coordinate (userBot) at ($(userC.south)+(0,-0.28)$);
\coordinate (merge3)  at ($(sysBot)!0.5!(userBot)$);
\draw[arr] (sysC.south)  -- (sysBot);
\draw[arr] (userC.south) -- (userBot);
\draw (sysBot) -- (merge3) -- (userBot);
\draw[arr] (merge3) -- ++(0,-0.3) -- ++(-4.4,0) -- ++(0,2.05) -- (s4h.east);

\coordinate (llmout) at ($(llmC.south)+(0,-0.28)$);
\draw (llmC.south) -- (llmout);
\draw[arr] (llmout) --  (s5h.north);
 
\coordinate (parseBot) at ($(parseC.south)+(0,-0.28)$);
\coordinate (saveBot)  at ($(saveC.south)+(0,-0.28)$);
\coordinate (merge5)   at ($(parseBot)!0.5!(saveBot)$);
\draw[arr] (parseC.south) -- (parseBot);
\draw[arr] (saveC.south)  -- (saveBot);
\draw (parseBot) -- (merge5) -- (saveBot);
\draw[arr] (merge5) -- ++(0,-0.3) -- ++(5,0) |- (artT.west);

\node[steplbl, below=13cm of s1h, xshift=-8cm] (legh) {Color legend:};
\node[sViolet, right=0.3cm of legh,  minimum width=2.0cm, text width=1.9cm, font=\scriptsize\bfseries] (l1) {Mode / instr.};
\node[sTeal,   right=0.2cm of l1,    minimum width=2.0cm, text width=1.9cm, font=\scriptsize\bfseries] (l2) {Topic / schema};
\node[sOrange, right=0.2cm of l2,    minimum width=2.0cm, text width=1.9cm, font=\scriptsize\bfseries] (l3) {Persona};
\node[sYellow, right=0.2cm of l3,    minimum width=2.0cm, text width=1.9cm, font=\scriptsize\bfseries] (l4) {System msg};
\node[sBlue,   right=0.2cm of l4,    minimum width=2.0cm, text width=1.9cm, font=\scriptsize\bfseries] (l5) {LLM call};
\node[sGreen,  right=0.2cm of l5,    minimum width=2.0cm, text width=1.9cm, font=\scriptsize\bfseries] (l6) {Parse / save};
 
\end{tikzpicture}
    \caption{Prompt construction pipeline for the Cognitive Digital Shadows dataset. Each step corresponds to the underlying Python implementation. Step~1 randomly selects a generation mode (\textit{human} or \textit{llm}),  a debate topic from a pool of four societal issues, and, if mode is  \textit{human}, a fully randomized 17-attribute persona via  \texttt{random\_persona()}.  Step~2 assembles the prompt inside \texttt{build\_prompt()}, combining  role-specific instructions, a JSON schema hint specifying the required output  fields (\textit{opinion}, \textit{reasoning\_summary}, \textit{tone},  \textit{citations}), and the serialized input values.  Step~3 wraps the assembled prompt into a two-message array consisting of a  system message enforcing JSON-only output and a user message carrying the  full prompt string.  Step~4 submits the message array to the LLM via \texttt{call\_llm()} (supporting both locally hosted models and API-based endpoints.),  \texttt{call\_llm()}, which attempts three response formats in order of  preference: \texttt{json\_schema}, \texttt{json\_object}, and plain text   with repair.  Step~5 parses and validates the raw response using  \texttt{safe\_json\_loads()}, enforcing the presence of all required fields,  and serializes the complete run artifact as a timestamped JSON file.  The color scheme follows the IBM colorblind-safe palette: violet for mode  and instructions, teal for topic and schema, orange for persona, yellow for  the system message, blue for the LLM call, and green for parsing and saving.}
    \label{fig:promptbuilding}
\end{figure}

The prompt pipeline described above (Fig. \ref{fig:promptbuilding}) provides a reproducible procedure for eliciting structured, opinion-style LLM responses under controlled role and persona conditions. Each design decision is grounded in peer-reviewed prompt engineering literature \cite{jacobsen2025promises,binz2023using,bender2021dangers} and validated LLM simulation frameworks \cite{Mahale2025,DeDuro2025}. Encapsulating these practices within a reproducible, structured pipeline is particularly important given that the intended end users of the generated data, domain experts such as psychologists, may not possess the prompt engineering competencies required to construct prompts of equivalent reliability independently to generate similar datasets.

\subsubsection*{Computational Requirements and Gap Filling}

The target audience for CDS is psychologists, cognitive scientists, communicators, computational social scientists, and political scientists, who need reusable data to audit discourses generated by LLMs on sensitive issues \cite{williams2026biased,DeDuro2025}. The appeal of Cognitive Digital Shadows mainly lies in two important aspects:

\begin{enumerate}
    \item the technical gap addressed by CDS, addressing long computational costs for mining LLM-generated data \cite{samsi2023words};
    \item user-friendly data access through CDS' data availability, enabling graphical user interfaces for semi-automated opinion polls and emotional analyses \cite{Semeraro2025}. 
\end{enumerate}

CDS bridges a technical gap mainly in terms of democratizing access to LLM-generated data. CDS helps reduce the technical access gap by releasing a large corpus of LLM outputs generated both locally and via API-based models, along with rich metadata and analysis-ready network representations. The debate generation process of CDS is rather computationally intensive and requires expensive technical infrastructure. The computational and methodological demands of the mining procedure are beyond the practical reach of most experts working without dedicated resources and support. Deploying and querying LLMs locally requires substantial hardware resources; in the present study, data generation was distributed across three dedicated machines (see Tab. \ref{tab:hardware}), with an average generation time of approximately 1,200 minutes per simulated model. 

\begin{table}[h!]
\caption{Hardware infrastructure used for data generation and post-processing. Runtime values are reported separately for LLM inference and TFMN feature extraction. For LLM inference, we report total wall-clock time and average time per valid record where available. Hardware names and GPU memory values correspond to the machines used during generation.}
\label{tab:hardware}
\centering
\renewcommand{\arraystretch}{1.3}
\footnotesize
\resizebox{\textwidth}{!}{
\begin{tabular}{|p{3cm}|p{3cm}|p{7cm}|p{3cm}|}
\toprule
\rowcolor[HTML]{E67E22}
\textcolor{white}{\textbf{Process}} &
\textcolor{white}{\textbf{Machine}} &
\textcolor{white}{\textbf{Machine Specifications}} &
\textcolor{white}{\textbf{Avg. Runtime (min)}} \\
\midrule

\rowcolor[HTML]{FEF0E7}
\multirow{3}{*}{LLM Data Generation}
& HP Z1 G9 Tower  
& GPU:  RTX 3060 12GB VRAM; CPU: Intel Core i7-13700, 16 cores; RAM: 32 GB;
& \multirow{3}{*}{ $\sim$29,280* (488 hours)} \\
\cline{2-3}

\rowcolor[HTML]{FEF0E7}
& Dell Alienware  Aurora R16 
& GPU: NVIDIA RTX 4090 (24 GB GDDR6X); CPU: Intel Core i7-14700F, 20 cores; RAM: 32 GB DDR5; Storage: 2 TB NVMe SSD 

& \\
\cline{2-3}

\rowcolor[HTML]{FEF0E7}
& Dell PowerEdge  R7525 
& GPU: NVIDIA L40 (48 GB); CPU: 2$\times$ AMD EPYC 7313, 32 cores total; RAM: 128 GB; Storage: 2$\times$960 GB SATA SSD 
& \\

\midrule

\multirow{3}{*}{\parbox{3cm}{Feature Extraction\\[4pt] TFMN Network Analysis \& Validation}}
& Galaxy Book 4 
& CPU: 13th Gen Intel Core i5 1335U; RAM: 16 GB; Storage: 1TB SSD; GPU: Iris(R)-Xe Graphics 
& $\sim$4380 \\
\cline{2-4}

& M75s-1 Desktop (ThinkCentre)
& CPU: AMD Ryzen 5 PRO 3400G (4 cores / 8 threads); RAM: 16 GB; Storage: SSD (NVMe);  GPU: AMD Radeon Vega 11; Windows 10 
& \multirow{2}{*}{$\sim$3600} \\
\cline{2-3}

& M75s-1 Desktop (ThinkCentre)
& CPU: AMD Ryzen 5 PRO 3400G(4 cores / 8 threads); RAM: 16 GB; Storage: 512GB SSD; GPU: AMD Radeon Vega 11; Windows 11 
& \\

\bottomrule
\end{tabular}
}

\vspace{0.6em}
\footnotesize
\textit{* Note.} $\sim$1,200 minutes per model across models; however, the most computationally intensive model (DeepSeek 70B) required approximately one week of continuous computation on a dedicated high-memory GPU system.

\end{table}

The most computationally intensive model, DeepSeek 70B, required approximately one week of continuous computation on a server with an NVIDIA L40 costing over 30K EUR. Feature extraction (comprising co-occurrence network construction and graph-theoretic analysis) was subsequently performed across three additional machines, with an average processing time of 360 minutes per run on average. CDS bridges this second gap by providing data generated from an expensive technical infrastructure to the general public and to researchers without funding support or access to such costly equipment (e.g. a server for simulating 70B LLMs can cost up to 30K EUR, \cite{guo2025deepseek}). 

In addition, CDS also provides a user-friendly pooling system, enabling researchers to perform complex queries in the database with a simple-to-use graphical user interface. Through the pooling system, without any need for coding, researchers can produce high-quality data visualisations, suitable for scientific publishing \cite{DeDuro2025}.

\subsubsection*{Data generation and processing}

Beyond hardware, the pipeline integrates expertise spanning prompt engineering \cite{jacobsen2025promises}, cognitive network science \cite{haim2026cognitive}, and natural language processing \cite{DeDuro2025,stella2020text}, disciplines that can fall outside the standard training of a single category of researchers interested in debate data, e.g. social scientists performing human-based coding of texts or psychologists running focus groups and/or psychometric questionnaires. 

The present framework was developed by an interdisciplinary team combining expertise in large language models, data science, data management, and network science, enabling the design and validation of a methodology that would be intractable for any single discipline to produce independently. By encapsulating the entire workflow within a reproducible, parameterized pipeline, the resulting resources provide psychological researchers with direct access to high-quality, validated synthetic debate data without requiring proficiency in the underlying computational and methodological disciplines.

\section*{Data Records}


Figure~\ref{fig:pipeline_generation} presents the 17 persona attributes randomly assigned in human-simulated generation mode, spanning 5 categories: biological, psychological, demographic, jobs and education, and personality and well-being. Two conditional dependencies govern the sampling: occupation is stratified by education level and migration status is conditioned on city of living, with an immigrant probability of $\approx 20.8\%$ derived directly from the generation script threshold. Fields in each CDS entry include $record\_id$, $model\_id$, $generation\_time$, $mode$, $topic$, $persona$, $opinion$, $reasoning\_summary$, $tone$, $citations$, $prompt\_config$, and $validation\_status$. The persona field includes 17 attributes in the human simulation mode but is assigned to null in the AI-assistant mode. All valid entries meet the schema validation criteria that require non-null mode, topic, opinion, and $reasoning\_summary$ fields and full persona data if mode is human.

Persona attributes were sampled according to predefined categorical or integer-valued probability distributions. Categorical variables were assumed to be sampled uniformly unless stated otherwise. Two conditions were introduced, where occupation was dependent on educational qualification and migration was dependent on the city of residence. All values, probabilities, and rules for conditional sampling can be found in Supplementary Table S1 and in the persona generation script that has been released. We did not attempt to match any real population distribution; the purpose of the sampling design was experimental coverage of prompt contexts rather than demographic representativeness.

\begin{figure}[ht!]
    \centering
    \includegraphics[width=\linewidth]{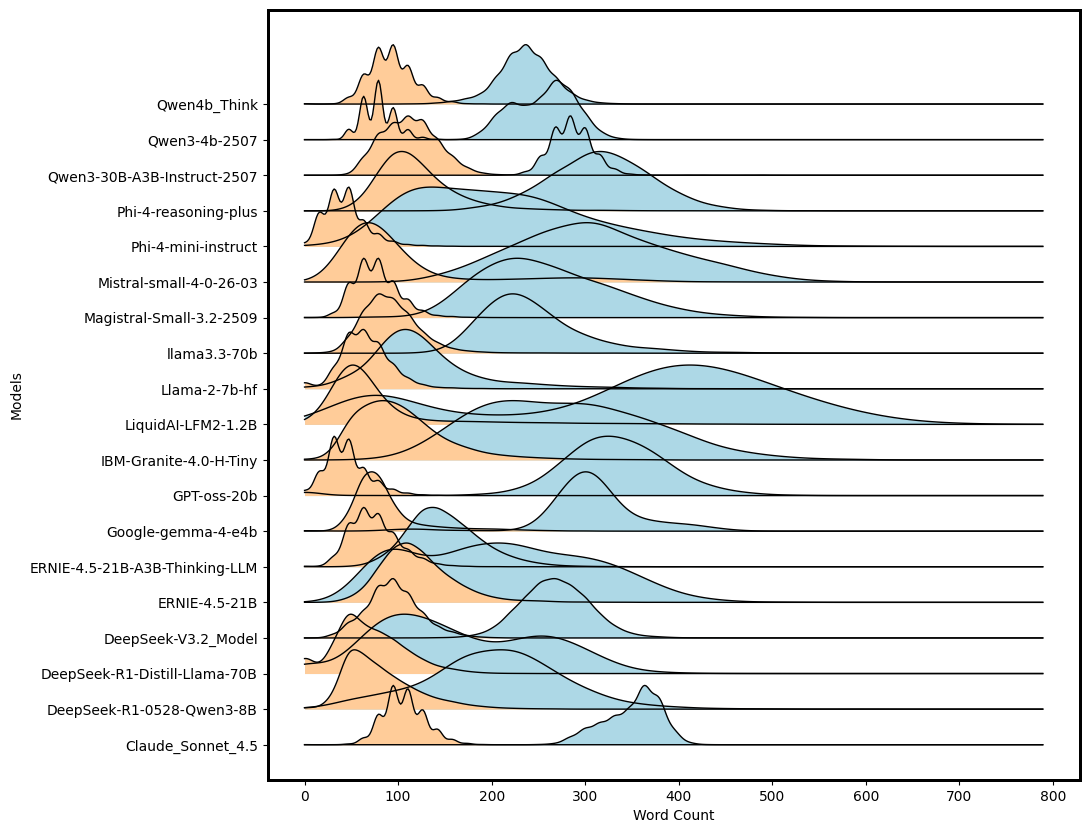}
    \caption{Distributions of word count across opinion and reasoning summary layers for all models in the corpus.   }
    \label{fig:wordcount_ridgeline}
\end{figure}

Figure~\ref{fig:wordcount_ridgeline} provides a corpus-level overview of word count distributions across all 19 models, separated by textual layer.

The ridgeline plot reveals that opinion texts consistently exhibit higher word counts than reasoning summaries across the majority of models, consistent with the prompt specification instructing continuous-prose opinions of 250--500 words alongside concise justifications. Notably, Qwen30b appears to be an exception to this pattern, with reasoning summaries exceeding opinion texts. This suggests that the model may produce comparatively more elaborate justification fields, This pattern should be interpreted together with the empirical word-count summaries provided in the repository.

Table~\ref{tab:llm_perspectives} presents two representative records from the generated corpus, illustrating two representative records from the four-topic corpus, \textit{COVID vaccine management}, \textit{fake content on social media}, \textit{gender gap in science}, and \textit{stereotype threat in STEM}, each decomposed into two textual layers: a discursive \textit{opinion} of 250--500 words and a concise \textit{reasoning summary} capturing the 
argumentative logic underlying the stated position.

For each generated debate text, the corresponding TFMN is constructed using EmoAtlas. The edge lists for both layers (opinion and reasoning summary) are stored as \texttt{parquet} files, which preserve the full network representation as structured records. Each file includes the complete edge list encoding syntactic and semantic relations between words, the degree sequence of all unique vertices representing their connectivity within the network, the occurrence frequency of each vertex across the corpus, and summary statistics such as the total number of nodes and edges. All network-level records are publicly available at \url{https://github.com/NaviDATA-Repos/PENSO_Data_WP-ConvinceMe_FIS2_UniTrento.git}.

\begin{table}[h!]
\caption{LLM Generated Perspectives and Persona. Two sample records illustrate the dataset structure across four controversial debate topics (\textit{COVID vaccine management}, \textit{fake content on social media}, \textit{gender gap in science}, and \textit{stereotype threat in STEM}), each decomposed into two textual layers: an \textit{opinion}, capturing the directly stated position; and a \textit{reasoning summary}, which is a structured justification of how the persona's sociodemographic and psychological attributes shaped the generated opinion. The first record is an LLM-generated (model: Qwen/Qwen3-4b-2507) entry without persona assignment. The second demonstrates human-simulated generation, where a fully specified sociodemographic persona (comprising age, biological sex, gender identity, sexual orientation, occupation, city of living, employment status, education level, parental education, marital status, number of children, migration status, psychological features, religious beliefs, social media usage, hobbies, and OCEAN personality scores) conditions the model toward a contextually grounded and demographically coherent perspective, yielding a qualitatively distinct rhetorical register from the persona-free condition.}
\label{tab:llm_perspectives}
\centering
\renewcommand{\arraystretch}{1.3}
\footnotesize
\resizebox{\textwidth}{!}{
\begin{tabular}{|p{1.8cm}|p{1.5cm}|p{2.5cm}|p{3.5cm}|p{3.5cm}|p{1.5cm}|p{1.5cm}|}
\toprule
\rowcolor[HTML]{E67E22}
\textcolor{white}{\textbf{Topic}} & 
\textcolor{white}{\textbf{Generation Mode}} & 
\textcolor{white}{\textbf{Persona Profile}} & 
\textcolor{white}{\textbf{Stated Opinion}} & 
\textcolor{white}{\textbf{Reasoning Summary}} & 
\textcolor{white}{\textbf{Tone}} & 
\textcolor{white}{\textbf{Model Used}} \\
\midrule
\rowcolor[HTML]{FEF0E7}
Gender gap in science: causes and solutions &
LLM &
Not applicable &
The gender gap in science is a complex phenomenon rooted in historical, social, 
and institutional dynamics rather than inherent ability. Addressing this gap is 
ethically imperative and strategically beneficial. ...
&
Analysis of historical context and institutional structures based on empirical 
evidence of underrepresentation. ...
&
Analytical, balanced, and solution-oriented &
qwen/ qwen3 -4b-2507 \\
\midrule
Management of fake content on social media &
Human-simulated &
33-year-old Native American female, Washington DC; bisexual, widowed, 1 child; 
unemployed retail worker with a bachelor's degree; Christian;  ...
&
Fake content deeply infiltrates daily life beyond viral posts or political 
debates. Simply ignoring false content is insufficient; there is a personal 
responsibility to engage cautiously. ...
&
Perspective rooted in personal experience and faith rather than technical 
expertise. The persona uses active questioning, source verification, and 
emotional regulation to manage misinformation. ...

&
Reflective, grounded, cautious, and personal &
qwen/ qwen3 -4b-2507 \\
\bottomrule
\end{tabular}
}
\end{table}

\section*{Technical Validation}

The validation analyses reported in this Section assess whether CDS' generated texts remain anchored to the prompted topic when represented as textual forma mentis networks \cite{haim2026cognitive,Semeraro2025}. These analyses do not validate the realism of persona simulation, the truth of generated claims, or the faithfulness of outputs to any human population \cite{liu2023trustworthy,de2025measuring}. Instead, they test a narrower property: Whether the semantic/syntactic structure of the generated texts is organized around topic-relevant concepts \cite{Semeraro2025}. We therefore interpret these analyses as topic-anchoring checks rather than as a validation of the realism of the CDS framework, which warrants more extensive research outside the scope of this Data Descriptor.

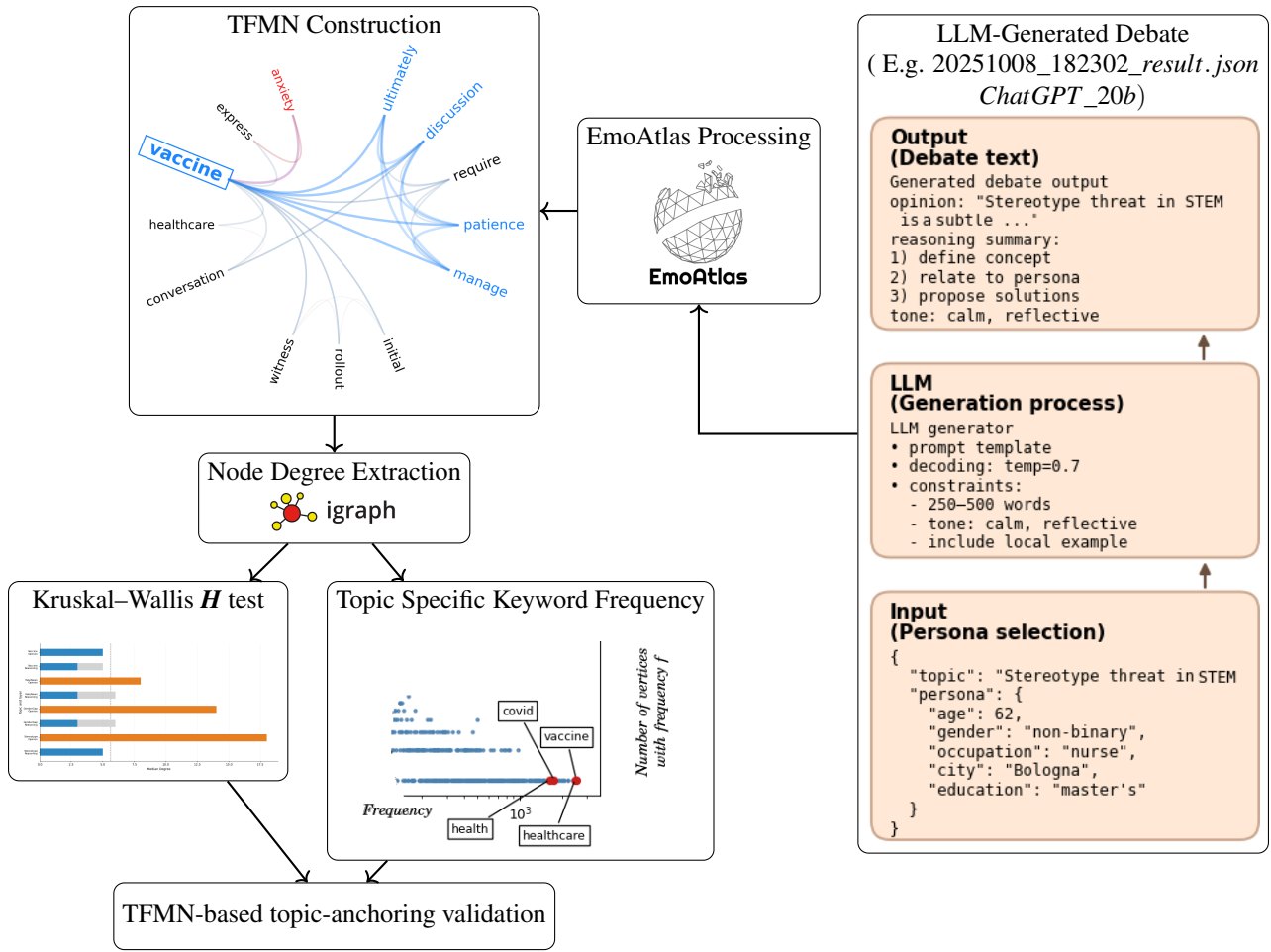
\begin{figure}[h!]
\centering
\begin{tikzpicture}[
    node distance=1.6cm,
    every node/.style={draw, rectangle, rounded corners, align=center, minimum width=3.2cm, minimum height=0.9cm},
    arrow/.style={->, thick},
    dashed arrow/.style={->, thick, dashed}
]
\node[anchor=north] (emo)
{EmoAtlas Processing \\ \includegraphics[width=0.1\linewidth]{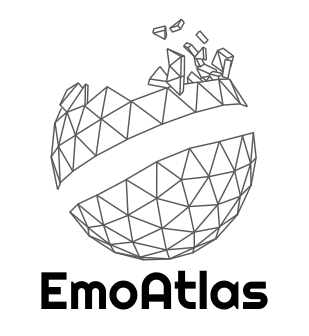}};

\node[anchor=north, right=0.5cm of emo, yshift=-3cm] (text) {LLM-Generated Debate\\ ( E.g. $20251008\_182302\_result.json$ \\ $ChatGPT\_20b)$\\
\includegraphics[width=0.3\linewidth]{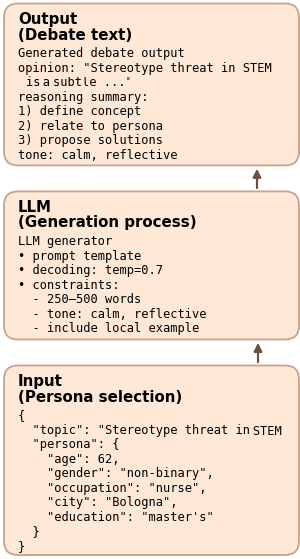} };

\node (tfmn) [left= 0.5 cm of emo] {TFMN Construction \\ \includegraphics[width=0.3\linewidth]{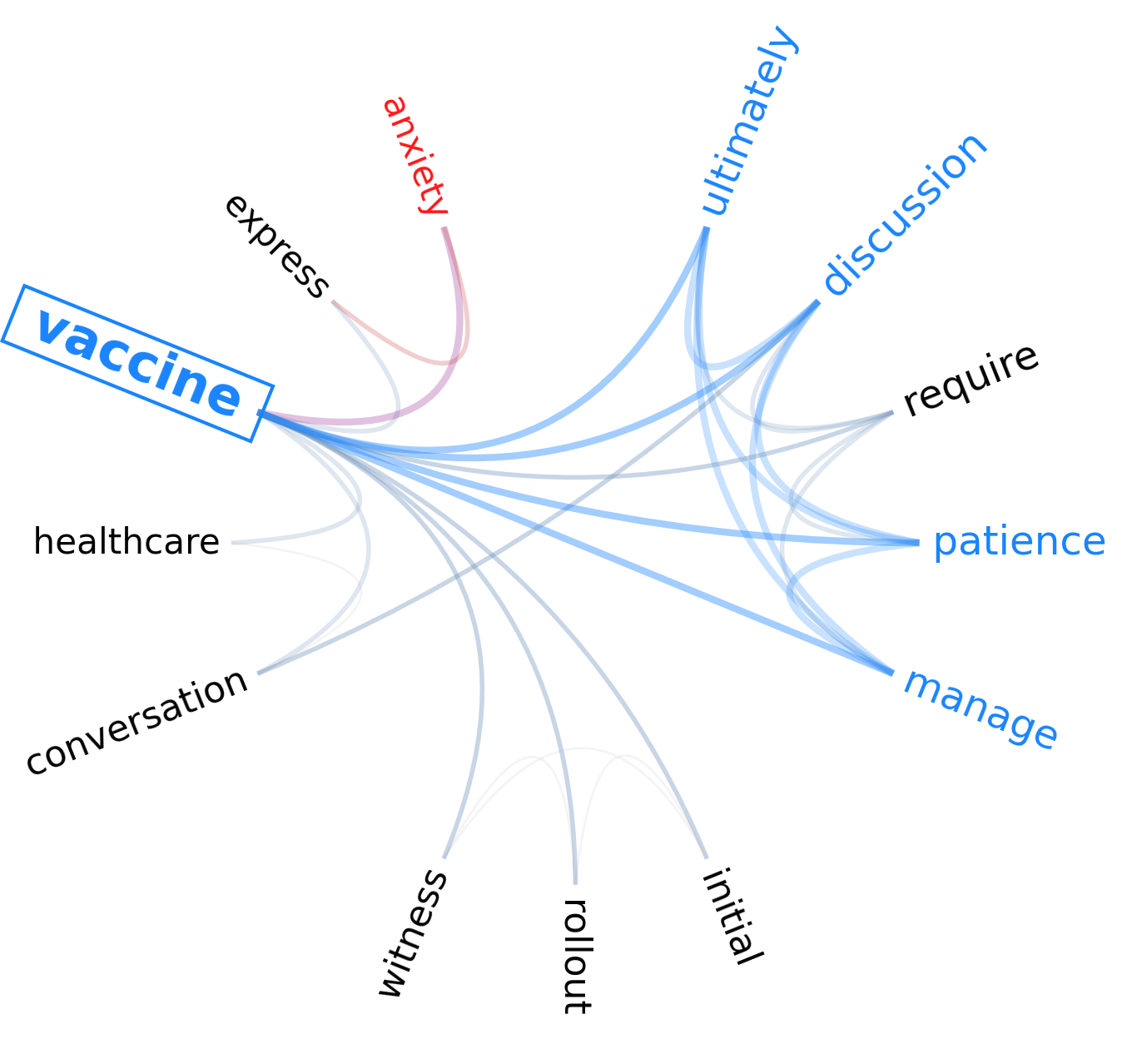}};

\node (deg) [below=0.5cm of tfmn]
{Node Degree Extraction\\\includegraphics[width=0.1\linewidth]{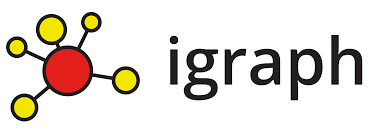}};

\node (perm) [below= 0.5 cm of deg, xshift=-2.5cm]  {Kruskal–Wallis \textit{\textbf{H}} test \\ \\ \includegraphics[width=0.2\linewidth]{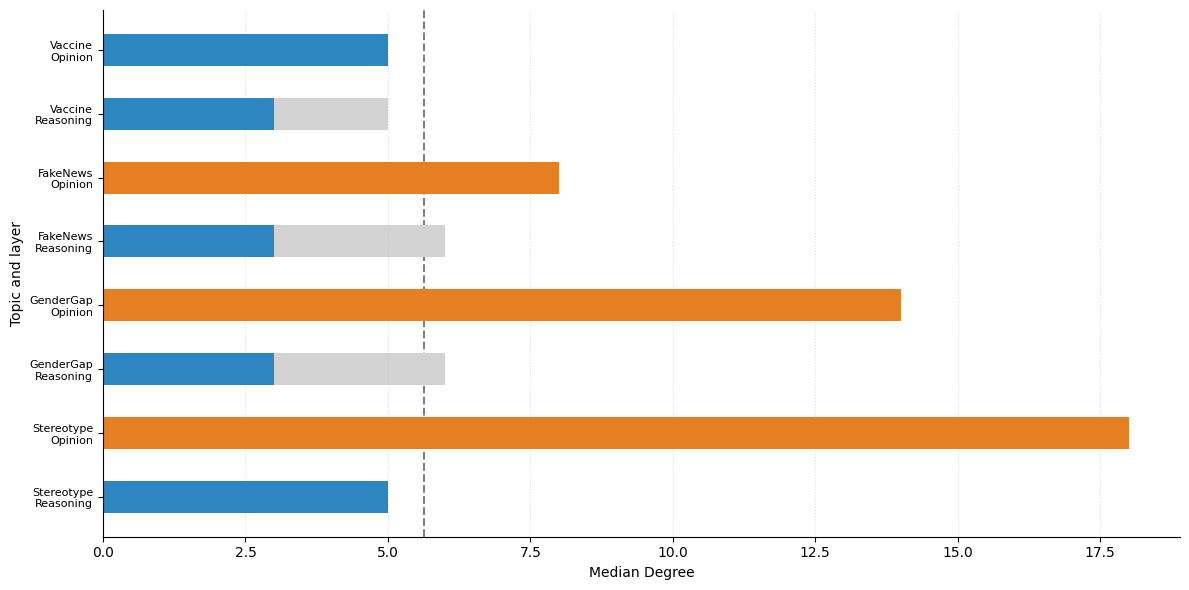}};

\node (Freq) [below= 0.5 cm of deg, xshift=2.5cm] {Topic Specific Keyword Frequency \\ \\ \includegraphics[width=0.25\linewidth]{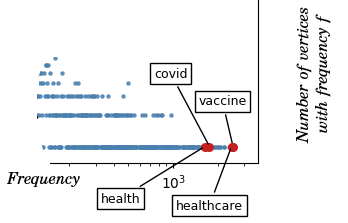}};

\node (val)  [below= 4.55 cm of deg] {TFMN-based topic-anchoring validation};

\draw[arrow] (text.west) -| (emo.south);

\draw[arrow] (emo)  -- (tfmn);
\draw[arrow] (tfmn) -- (deg);
\draw[arrow] (deg)  -- (perm);
\draw[arrow] (deg)  -- (Freq);
\draw[arrow] (perm) -- (val);
\draw[arrow] (Freq) -- (val);

\end{tikzpicture}
\caption{Analytical pipeline for validating textual forma mentis networks (TFMNs). The pipeline begins with LLM-Generated Debate texts, which serve as the raw corpus feeding into the analysis. These are processed through EmoAtlas, which extracts syntactic/semantic word associations from the text. The resulting data is then used to construct a TFMN, where words are linked and clustered, forming a structured representation of conceptual thought. From this network, Node Degree Extraction identifies how central each word is within the semantic structure. This centrality measure then branches into two parallel validation paths: a Kruskal–Wallis H test, which statistically compares degree distributions across different groups to detect meaningful structural differences, and Topic-Specific Keyword Frequency analysis, which grounds the network patterns in direct lexical evidence.}
\label{fig:pipeline}
\end{figure}

This validation assesses topic anchoring by testing whether topic-related concepts, such as “gender” and “gap,” occupy more central positions in the extracted TFMNs than comparison terms. In other words, we want to measure whether the LLMs tend to speak more commonly and with increased syntactic/semantic richness about concepts expressing the key topics of conversations rather than other words. The core hypothesis is that, if LLM-generated texts reliably encode the prompted topic and produce a coherent topic-focused discourse, then keywords that are syntactically/semantically related to the topic should emerge as structurally central nodes in the corresponding TFMNs. To operationalize topic anchoring, we measured node degree $k_i$ for topic-related keywords and for all other non-stopword, non-topic vertices in each TFMN. We then compared the two degree distributions using a non-parametric rank-based test. Because only two groups were compared, we report the Kruskal–Wallis statistic as equivalent to a two-group rank comparison. The full analytical pipeline is illustrated in Fig.~\ref{fig:pipeline} and described step by step below.

Let $C \subset V$ denote the set of topic-related keywords derived from the original LLM prompt and mapped to the TFMN vocabulary. In Step~4 of Fig.~\ref{fig:pipeline}, we define the observed keyword degree statistics as vertex centrality \(C_{\text{kwd}}\)\cite{igraph-2} using igraph \cite{igraph}. Given that the degree distribution is not normal, and the values are analyzed in terms of their ranks, the statistical significance of keyword centrality is evaluated using the nonparametric, Kruskal–Wallis test \cite{Kruskal1952,Gauthier2015}. In this stage, we compare \textbf{List~1} (the degree values of all non--topic-specific vertices after filtering) against \textbf{List~2} (the degree values of the $C_{\text{kwd}}$). Accordingly, the null hypothesis is formulated as follows:
\textit{$H_0$: The degree values of topic-keyword vertices and comparison vertices are drawn from the same rank distribution.} Rejection of $H_0$ indicates a difference in degree distributions under the adopted TFMN representation; it does not, by itself, establish discourse quality, factual accuracy, or persona realism, which are left as research results for future studies.

Importantly, as each generated text was related to one specific topic and each validation test to a different set of records, we interpreted each statistic as a topic-anchoring check for a specific model, topic, and textual component instead of testing one set of hypotheses for a family of validation tests. Thus, for all comparisons, we report test statistics and P values separately, without performing multiple comparisons adjustments.

Under the null hypothesis of identical distributions of degrees, $H$ is used to test whether the two groups differ significantly \cite{scipy_reference}. For each topic and textual layer, we compare the degree distribution of topic keywords with that of non-topic, non-stopword vertices. Because the degree distributions are strongly skewed, we use a nonparametric group comparison. We do not interpret rejection of the null hypothesis as evidence of general discourse quality. It only indicates that the compared degree distributions differ under the adopted representation. Where needed, follow-up comparisons are reported separately.





Across the tested model-topic-layer combinations, the resulting $H$ statistics indicate differences between the degree distributions of topic-keyword vertices and comparison vertices (Fig. 5). These results support the interpretation that generated texts are structurally anchored to the prompted topics under the adopted TFMN representation.

\begin{figure}[h!]
    \centering
    \includegraphics[width=0.93\linewidth]{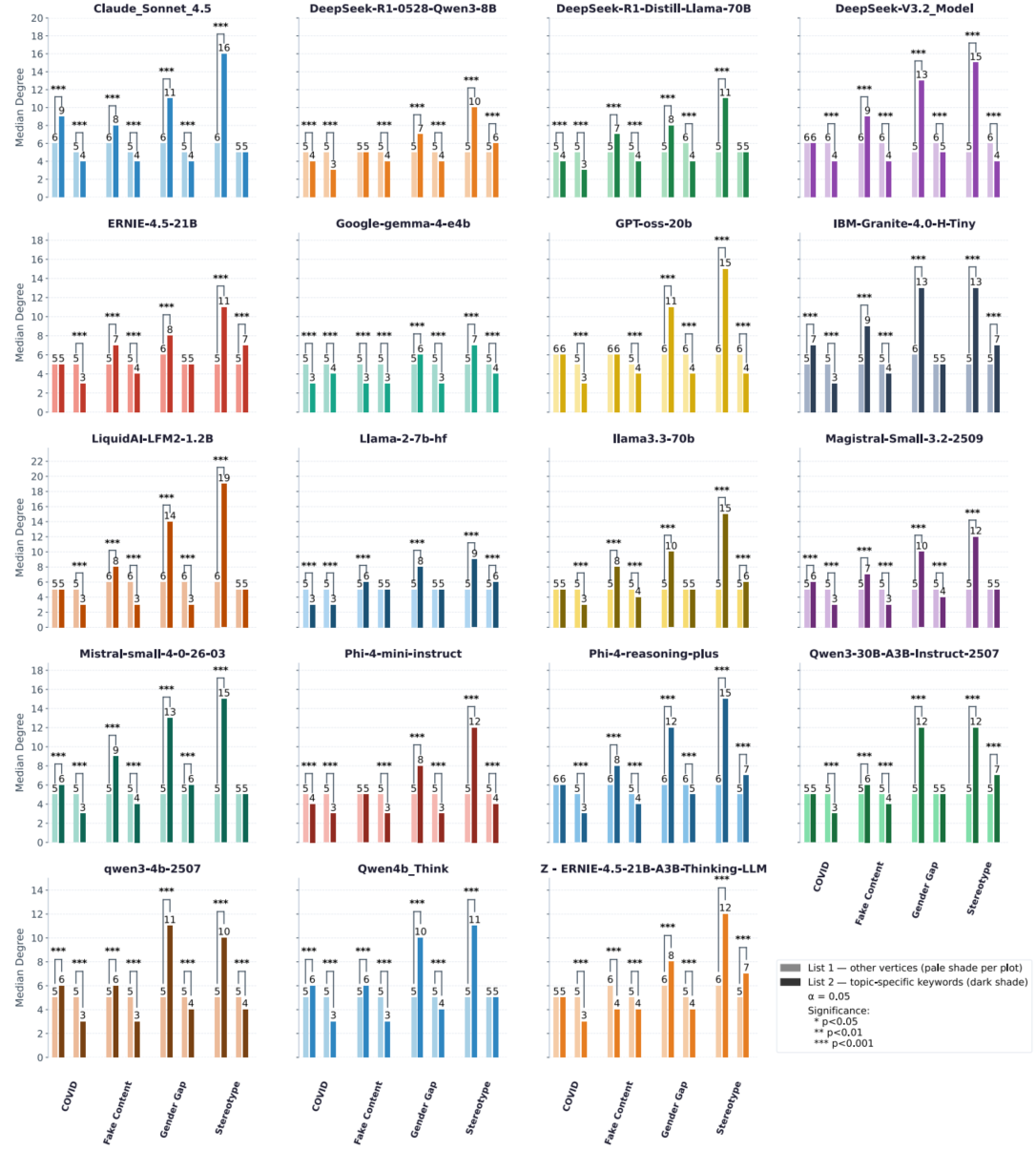}
    \caption{Kruskal–Wallis test results comparing the node degree of $C_{\text{kwd}}$ against all other vertices across textual forma mentis networks (TFMNs) extracted from debates generated by 19 LLMs on four controversial topics (\textit{Vaccine-Covid}, \textit{FakeNews}, \textit{Gender Gap}, \textit{Stereotype-STEM}); each analyzed at two layers (\textit{Opinion} and \textit{Reasoning\_Summary}). For each topic-layer combination, the node degree (a measure of conceptual centrality within the TFMN) is compared between $C_{\text{kwd}}$ (\textit{list2}, dark shade) and the remaining non-stopword, non-topic-word vertices (\textit{list1}, pale shade) using a non-parametric test. Statistical comparisons were conducted using the Kruskal–Wallis rank-sum test via \texttt{scipy.stats.kruskal}. 
    Each subplot corresponds to one LLM, with bar color uniquely assigned per model. For each topic, two bar pairs are shown: the left pair represents the Opinion layer and the right pair represents the Reasoning Summary layer.  The H-statistic is reported below each topic-layer pair. Subplots within the same row share a common y-axis scale. All tests were conducted at significance level $\alpha = 0.05$.
    }
\label{fig:kruskal_wallis}
\end{figure}

However, for the reasoning summary layer, although the Kruskal--Wallis test statistic $H$ indicates a statistically significant difference and thus leads to rejection of the null hypothesis, the median of all $C_{\text{kwd}}$s pooled together (not per keyword, since they are all topic-specific keywords, all their degrees flattened into one list in this analysis) is lower than that of the remaining words in the corpus. Therefore, we further compare each $C_{\text{kwd}}$ with the other words within its respective topic (excluding stopwords and topic words) across both the opinion and reasoning summary layers.

\begin{figure}[h!]
    \centering
    \includegraphics[width=0.7\linewidth]{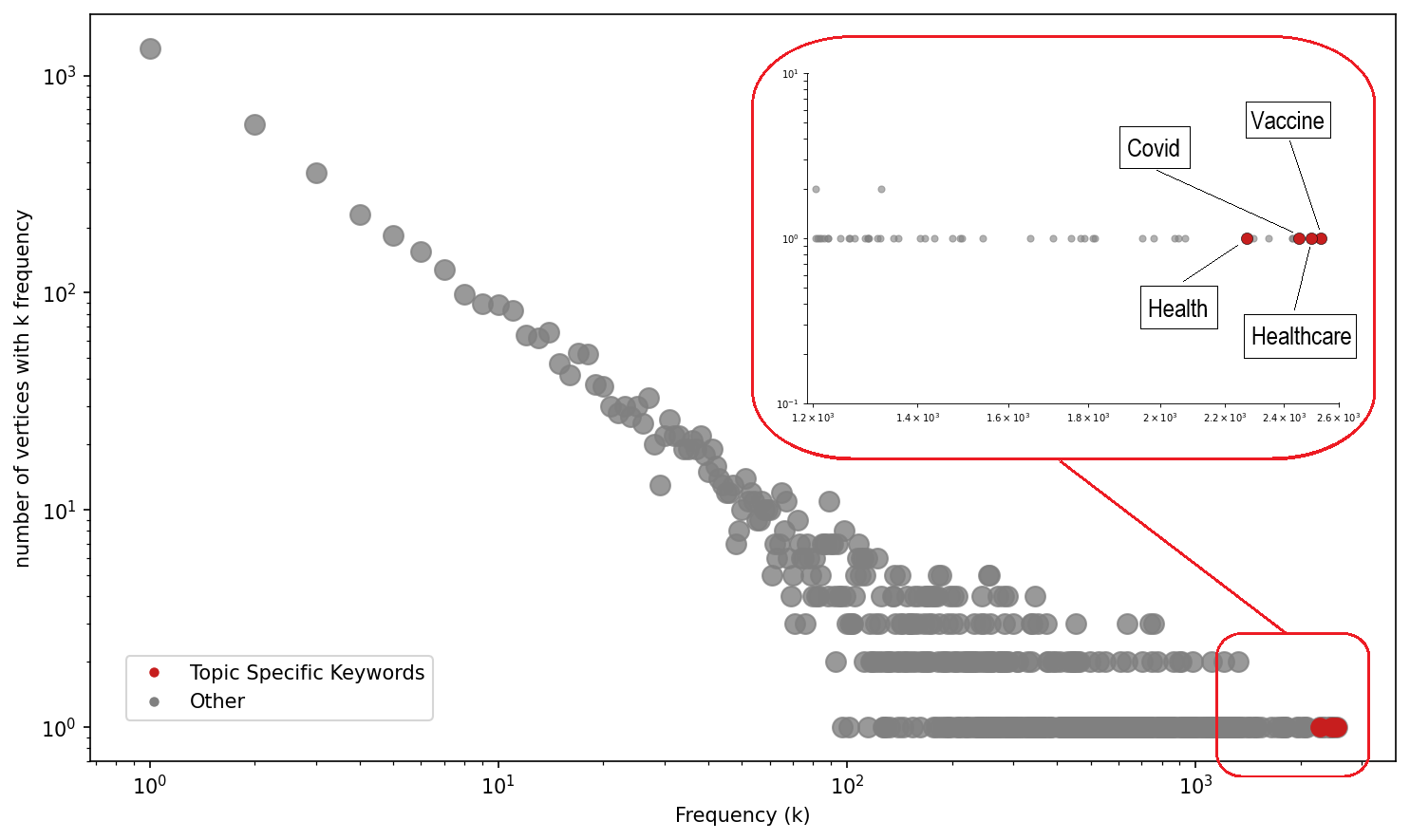}
\caption{Log-log frequency distribution of vertex occurrences across the TFMN 
extracted from Claude Sonnet 4.5 generated debates on the \textit{Vaccine-Covid} topic. Each point represents the number of vertices (y-axis) sharing a given occurrence frequency $f$ (x-axis). The distribution follows a characteristic power-law decay, where the vast majority of words appear rarely while only a small subset recurs with high frequency. $C_{\text{kwd}}$ including \textit{covid}, \textit{vaccine}, \textit{health}, and \textit{healthcare} are highlighted in red. $C_{\text{kwd}}$ are concentrated at the rightmost tail of the distribution, indicating that they are among the most frequently occurring terms in the generated text. This result provides evidence that the generated debates are semantically anchored around the intended concepts.}
\label{fig:freq_dist}

\end{figure}

The other validation method in Step~5 of Fig.~\ref{fig:pipeline}, is used to study the occurrence of $C_{\text{kwd}}$, verifying that they appear among the highest-frequency vertices in the TFMN. The results in Fig. \ref{fig:freq_dist} illustrates that the LLM-generated discourse is lexically dominated by the expected conceptual anchors rather than peripheral or incidental terms.

Degree centrality captures syntactic/semantic connectivity rather than mere repetition. Statistically significant keyword centrality therefore constitutes evidence that the LLM-generated text embeds the prompted topic in a coherent conceptual structure. On the other hand, occurrence frequencies of $C_{\text{kwd}}$ confirm the expected structural and lexical dominance of $C_{\text{kwd}}$. As summarized in Step~6 of Fig.~\ref{fig:pipeline}, statistically significant keyword centrality provides evidence that the LLM-generated text embeds the prompted topic in a coherent syntactic/semantic structure rather than producing semantically diffuse content.



\section*{Usage Notes}


The Cognitive Digital Shadows dataset can be reused for comparisons between responses generated by different language models, on different topics, different role modes, and different attributes of persona-prompt combinations. Recommended use cases for CDS reuse include descriptive audits of the framing, affective tone, topic grounding, and sensitivity to the attributes conditioned on prompts. Users are cautioned against interpreting CDS data as actual measures of human beliefs, attitudes of human demographic groups, or human psychological states. Nonetheless, CDS can power future comparisons between LLM- and human data matching specific demographics or psychological features.


\subsection*{Textual forma mentis network construction}

As shown in Fig.~\ref{fig:pipeline}, each generated text constitutes the input document for subsequent network extraction. Each text is processed using EmoAtlas to construct a textual forma mentis network \cite{haim2026cognitive,Semeraro2025,stella2020text}. In a TFMN, nodes represent concepts/words/lexical items (e.g. "gender", "gap", "science", etc.) linked by semantic overlap relationships (i.e. synonyms, like "gap" and "divide") or by syntactic specifications (i.e. "love" is "weakness"). Following prior TFMN research, we use the terms “concepts,” “words,” and “lexical items” to refer to the lemmatized nodes represented in the network \cite{haim2026cognitive,abramski2023cognitive,stella2020text}. In mathematical terms, a TFMN can be identified as a network $\mathcal{N}:=(V,E)$, where $V$ is the set of lemmatized non-stopwords (i.e. nodes or words), while $E$ is the set of syntactic-semantic links between words.

\begin{figure}[h!]
\centering
\begin{tikzpicture}[
    every node/.style={inner sep=0pt},
    arrow/.style={->, line width=3pt, gray!70}
]

\node (A) {\includegraphics[width=0.5\linewidth]{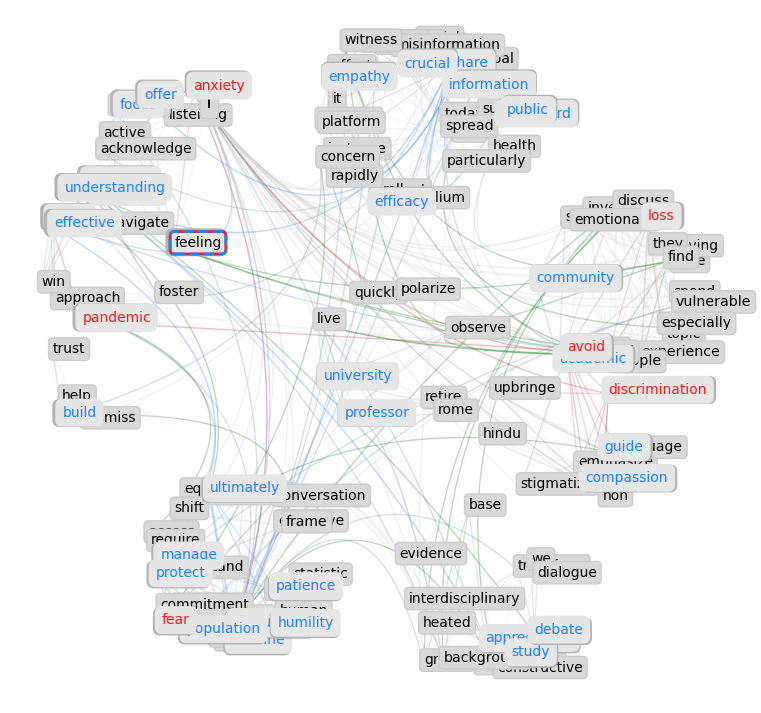}};
\node[anchor=north west, font=\bfseries\Large, text=gray!70] at (A.north west) {A};

\node (B) [right=0.5cm of A] {\includegraphics[width=0.5\linewidth]{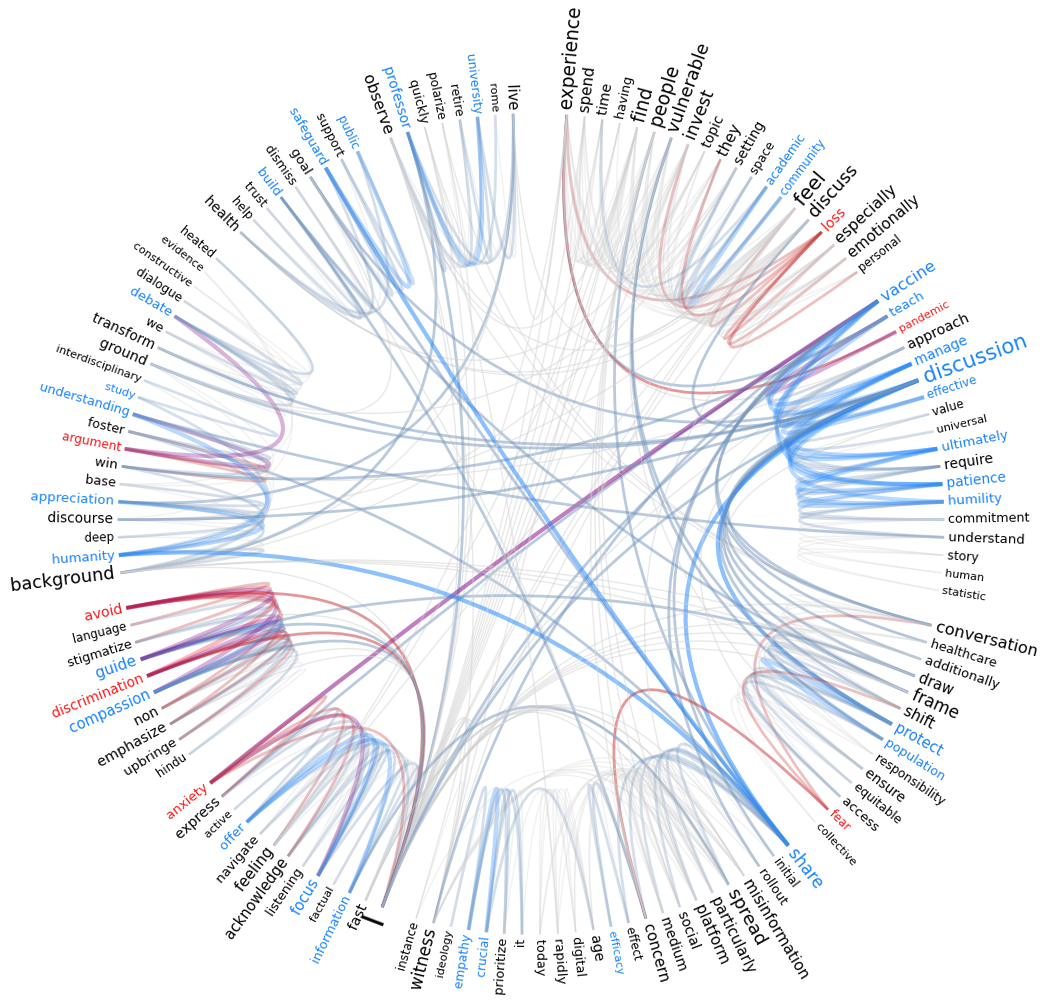}};
\node[anchor=north west, font=\bfseries\Large, text=gray!70] at (B.north west) {B};

\node (C1) [below=1.2cm of B, xshift=-0.3\linewidth] {\includegraphics[width=0.3\linewidth]{figures/vaccine.png}};
\node[anchor=north west, font=\bfseries\Large, text=gray!70] at (C1.north west) {C1};

\node (C2) [right=0.4cm of C1] {
\resizebox{6.5cm}{!}{
\begin{tikzpicture}    [bluenode/.style={draw=nodeblue, fill=none, text=nodeblue, rounded corners=3pt, inner sep=2pt, font=\small},
    rednode/.style={draw=nodered, fill=none, text=nodered, rounded corners=3pt, inner sep=2pt, font=\small},
    graynode/.style={draw=none, fill=none, text=nodegray, inner sep=2pt, font=\scriptsize},
    blueedge/.style={draw=edgeblue, line width=1.8pt, opacity=0.85},
    rededge/.style={draw=edgered, line width=1.8pt, opacity=0.85},
    purpedge/.style={draw=edgepurple, line width=1.8pt, opacity=0.85},
    grayedge/.style={draw=edgegray, line width=1.4pt, opacity=0.75},
    bluefaint/.style={draw=edgeblue, line width=1.0pt, opacity=0.35},
    redfaint/.style={draw=edgered, line width=1.0pt, opacity=0.35},
    grayfaint/.style={draw=edgegray, line width=1.0pt, opacity=0.40},
    purpfaint/.style={draw=edgepurple, line width=1.0pt, opacity=0.35},
]
\definecolor{nodeblue}{RGB}{70,145,190}
\definecolor{nodered}{RGB}{200,80,80}
\definecolor{nodegray}{RGB}{160,160,160}
\definecolor{edgeblue}{RGB}{100,160,210}
\definecolor{edgered}{RGB}{210,130,130}
\definecolor{edgegray}{RGB}{170,170,170}
\definecolor{edgepurple}{RGB}{160,120,180}

\node[rednode] (pandemic) at (0, 0) {pandemic};
\node[bluenode] (vaccine) at (7, 0) {vaccine};
\node[bluenode] (teach) at (1.8, 0.7) {teach};
\node[graynode] (experience) at (1.8, -0.4) {experience};
\node[graynode] (find) at (3.2, 2.2) {find};
\node[graynode] (feel) at (3.2, 1.5) {feel};
\node[graynode] (I) at (3.4, 0.5) {I};
\node[graynode] (approach) at (3.1, -0.1) {approach};
\node[bluenode] (effective) at (3.2, -0.9) {effective};
\node[graynode] (discuss) at (3.2, -1.6) {discuss};
\node[graynode] (feeling) at (3.2, -2.3) {feeling};
\node[bluenode] (manage) at (5.2, 2.2) {manage};
\node[graynode] (express) at (5.2, 1.4) {express};
\node[graynode] (witness) at (5.2, 0.7) {witness};
\node[rednode] (anxiety) at (5.1, 0) {anxiety};
\node[graynode] (conversation) at (5.2, -0.9) {conversation};
\node[bluenode] (discussion) at (5.2, -1.6) {discussion};
\node[graynode] (rollout) at (5.2, -2.2) {rollout};

\draw[purpedge] (pandemic) -- (teach);
\draw[redfaint] (pandemic) -- (experience);
\draw[bluefaint] (discussion) -- (approach);
\draw[purpedge] (anxiety) -- (vaccine);
\draw[blueedge] (vaccine) -- (manage);
\draw[blueedge] (vaccine) -- (discussion);
\draw[blueedge] (vaccine) -- (witness);
\draw[bluefaint] (vaccine) -- (express);
\draw[bluefaint] (vaccine) -- (conversation);
\draw[bluefaint] (vaccine) -- (rollout);
\draw[bluefaint] (teach) -- (approach);
\draw[blueedge] (teach) -- (effective);
\draw[bluefaint] (teach) -- (I);
\draw[blueedge] (manage) -- (effective);
\draw[grayedge] (express) -- (I);
\draw[grayedge] (witness) -- (find);
\draw[grayedge] (witness) -- (feel);
\draw[bluefaint] (manage) -- (approach);
\draw[grayedge] (feeling) -- (express);
\draw[bluefaint] (I) -- (discussion);
\draw[grayedge] (I) -- (rollout);
\draw[grayedge] (I) -- (experience);
\draw[grayedge] (I) -- (witness);
\draw[grayedge] (I) -- (conversation);
\draw[grayedge] (experience) -- (find);
\draw[grayedge] (experience) -- (feel);
\draw[grayedge] (experience) -- (approach);
\draw[grayedge] (experience) -- (discuss);
\draw[grayedge] (experience) -- (feeling);
\draw[bluefaint] (discussion) -- (discuss);
\draw[redfaint] (I) -- (anxiety);
\draw[redfaint] (feeling) -- (anxiety);
\end{tikzpicture}
}
};

\node[anchor=north west, font=\bfseries\Large, text=gray!70] at (C2.north west) {C2};

\node (D) [below=1.3cm of A, xshift=-1.7cm] {\includegraphics[width=0.3\linewidth]{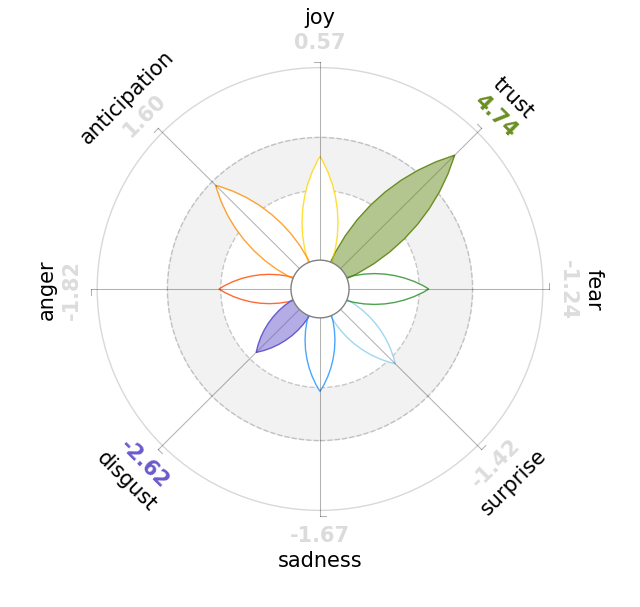}};
\node[anchor=north west, font=\bfseries\Large, text=gray!70]
    at (D.north west) {D};
\draw[arrow] (A) -- (B);
\draw[arrow] (B) -- (C1);
\draw[arrow] (B) -- (C2);

\end{tikzpicture}
\caption{A: The force-directed graph sample that shows a syntactic/semantic network for a debate (opinion) generated on the topic “How to manage and deal with discussions about healthcare and COVID vaccines.” B: Textual forma mentis network in the figure is revealing the conceptual-affective structure of the corpus. Nodes represent frequently occurring vertices arranged radially, with edges indicating syntactic/semantic relations colored by emotional valence (blue = positive, red = negative, purple = mixed/neutral). Edge thickness reflects association strength. C1: Ego network of the node \textit{vaccine}, showing its direct semantic and affective associations within the TFMN. C2: Mindset stream of \textit{vaccine} and \textit{pandemic} as anchor nodes. The presence of \textit{anxiety} as a shared neighbour further suggests that the generated text acknowledges the emotional dimension of the topic, positioning vaccine discourse not merely as a scientific matter but as an affectively loaded social negotiation. D: The emotional flower reveals a strongly trust-centered configuration (z = 4.74), significantly exceeding random baseline expectations (in gray). In contrast, all negative emotions, disgust, fear, anger, sadness remain below significance thresholds.}
\label{fig:TFMN}
\end{figure}

For instance, considering a data record in which the system is queried for texts discussing healthcare and COVID-19 vaccines among Rome residents holding a master's degree, biological sex male, and religion Hinduism, across all human modes. Under these sociodemographic constraints, TFMN of the corresponding persona corpus is constructed using the edge list in the dataset, and visualized in Fig.~\ref{fig:TFMN}.


\subsection*{Interpretable natural language processing with TFMNs}

TFMNs are interpretable network representations because each node and edge has a direct linguistic counterpart \cite{haim2026cognitive,stella2020text}: nodes correspond to lexical items, and edges correspond to syntactic or semantic relations extracted from the text. Therefore, TFMNs constitute a glass box framework for natural language processing, in which every element of the model is directly readable as a meaningful linguistic representation of the source text \cite{Semeraro2025}.

In TFMNs, node degree $k_i$ is defined as the number of links a given node $i$ participates in, as in network science. However, given the syntactic/semantic nature of TFMNs' links, $k_i$ can be interpreted as the semantic richness of individual words, i.e. the number of associations attributed to a given concept $i$ within a portion of text.
For instance, in the example of the text discussing the topic of dealing with discussions about healthcare and COVID vaccines in Fig. \ref{fig:TFMN}.

Since \textit{vaccine} is a topic-specific keyword ($C_{\text{kwd}}$) in this example, it is possible to extract and analyze  its ego network within the TFMN, revealing the syntactic/semantic relations that the LLM-generated discourse constructs around this central concept. The resulting ego network in Fig.~\ref{fig:TFMN}-C1 is overwhelmingly blue, indicating that \textit{vaccine} is predominantly associated with positively valenced 
concepts. 

One of the strongest connections, \textit{patience}, suggests a discourse emphasizing careful and measured debate, whereas \textit{manage} frames vaccines as instruments of public health control. The presence of \textit{discussion} and \textit{ultimately} points to rational, deliberative reasoning. The relative isolation of \textit{anxiety} as the sole strongly negative node suggests that vaccine hesitancy is acknowledged but not deeply engaged with, remaining peripheral to the dominant positive framing.

Another usage example is studying the mindset stream between key concepts within TFMN because it can reveal different cognitive orientations \cite{brian2023introducing}. The mindset stream between “vaccine” and “pandemic” can be used to inspect whether the generated discourse links the public-health problem and its proposed remedy through shared semantic associations. Fig. \ref{fig:TFMN}-C2, extracted from the TFMN in Fig.~\ref{fig:TFMN}, indicates that among the matched sociodemographic profile, two words are cognitively bridged through a shared cluster of experiential (\textit{experience}, \textit{discuss}, \textit{conversation}, and \textit{feeling}), and educational concepts (\textit{teach}, \textit{effective}, \textit{manage}) suggesting that the LLM-generated discourse frames both concepts primarily through the lens of personal and social dialogue rather than clinical or epidemiological reasoning. 

Notably, \textit{vaccine} is strongly associated with action-oriented terms such as \textit{manage}, and \textit{discussion}, reflecting a solution-driven discourse; whereas \textit{anxiety}, evoking a more emotionally charged orientation.

\subsection*{Dashboard for Data Pooling}

\begin{figure}[h!]
    \centering

        \includegraphics[width=\textwidth]{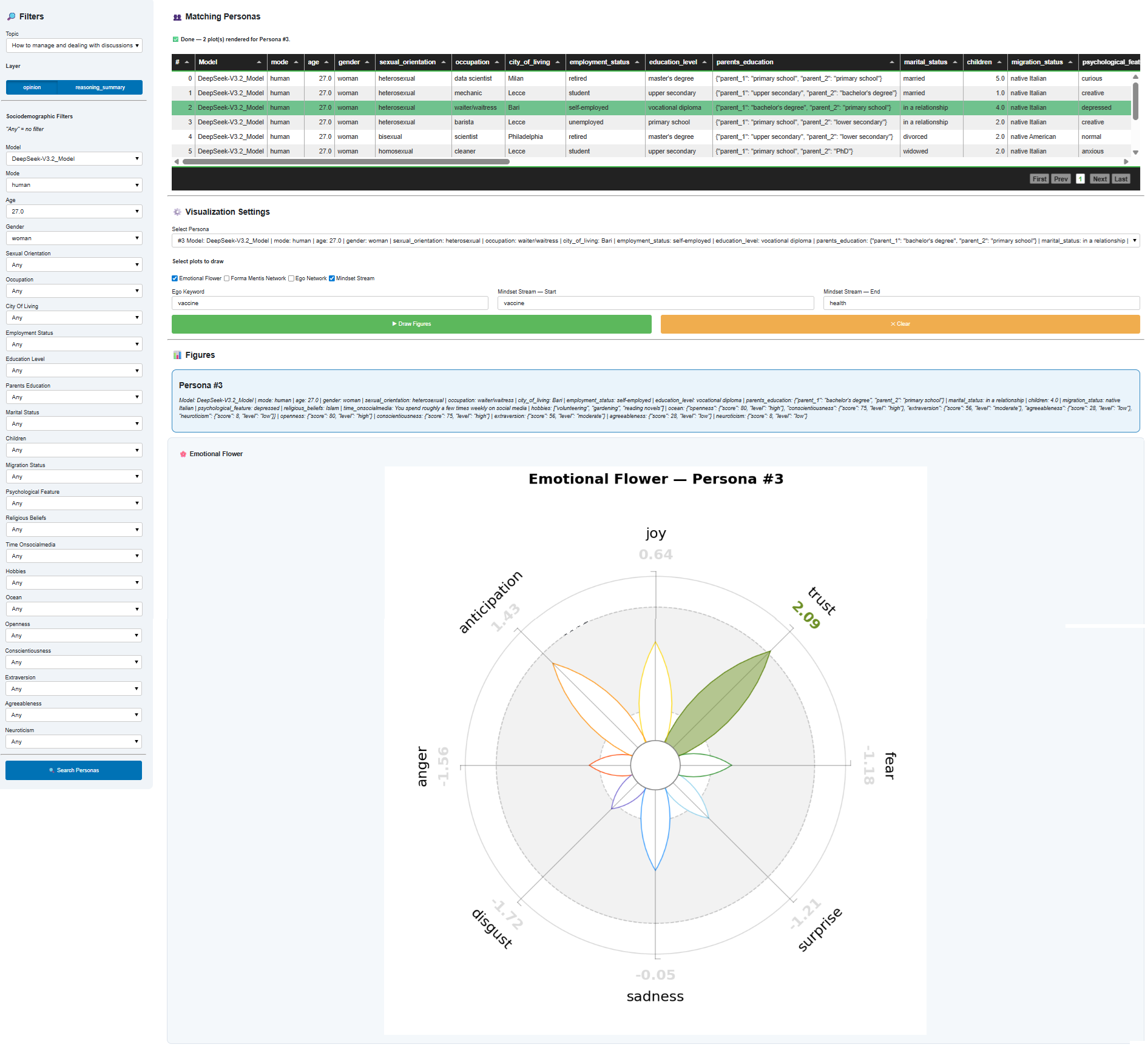}
        \caption{The left sidebar exposes the full set of filter controls: topic selector, textual layer toggle (\textit{opinion} / \textit{reasoning\_summary}), and sociodemographic dropdowns (biological sex, gender, sexual orientation, occupation, city of living, employment status, education level, marital status, migration status, psychological feature, religious beliefs, time on social media, and interaction mode), all set to \texttt{Any} by default. The main panel displays the empty \textit{Matching Personas} section and the \textit{Visualization Settings} panel, where the user can pre\-configure the ego keyword, mindset stream endpoints, and the subset of plots to render. The \textit{Draw Figures} button remains disabled until a search is executed. Persona retrieval is performed after applying sociodemographic filters. The user has set \textit{Biological Sex} to \texttt{male}, \textit{Gender} to \texttt{man}, \textit{City of Living} to \texttt{Bari}, \textit{Employment Status} to \texttt{student}, and \textit{Religious Beliefs} to \texttt{Christianity}, while selecting the topic \textit{How to manage and deal with discussions about healthcare and COVID vaccines}. Clicking \textit{Search Personas} queries the unified dataset and returns 55 matching personas, whose full sociodemographic profiles are displayed in a scrollable table. The \textit{Select Persona} dropdown is automatically populated with labeled entries encoding each persona's complete attribute set (in this case persona number 3 is selected), and the \textit{Draw Figures} button becomes active (in this case only the emotional flower is selected).}
        \label{fig:dashboard}
    \end{figure}

The interactive dashboard was developed in Google Colaboratory (Colab) (see Fig. \ref{fig:dashboard}) and provides a graphical interface for exploring the cognitive and emotional structure of textual forma mentis networks extracted from LLM-generated persona debates \cite{haim2026cognitive,Semeraro2025,stella2020text}. The dashboard leverages the \texttt{emoatlas} Python library~\cite{emoatlas} for network construction and visualization, and \texttt{ipywidgets} for interactive controls. Prior to launching the dashboard, the user must mount their Google Drive and provide the path to the unified dataset file (\texttt{MASTER\_parquet}), which contains all persona-level edge lists alongside their associated sociodemographic attributes. The dashboard operates in four sequential steps.

First, the user selects the debate topic of interest from a dropdown menu — choosing among the four controversial topics included in the dataset (\textit{Vaccine-Covid}, \textit{FakeNews}, \textit{Gender Gap in Science}, and \textit{Stereotype Threat in STEM}) — and specifies the textual layer to analyze, either \textit{Opinion} or \textit{Reasoning\_Summary}, corresponding to the two layers of the TFMN. Then, the user applies sociodemographic filters to narrow the persona pool. 

Available filters include biological sex, gender identity, sexual orientation, occupation, city of residence, employment status, education level, marital status, migration status, psychological profile, religious beliefs, time spent on social media, and interaction mode (human role-play vs.\ LLM assistant). Each filter defaults to \texttt{Any}, meaning no restriction is applied, and multiple filters can be combined simultaneously to isolate specific subpopulations of interest. 

Then, the user clicks the \textit{Search Personas} button, which queries the dataset and displays a structured table listing all personas that satisfy the selected filter combination, with their full sociodemographic profiles shown for inspection. The matching personas are also populated into a dropdown selector. Next, the user selects a single persona from the dropdown, chooses which visualisations to render via a multi-select widget — from among the \textit{Emotional Flower} \cite{Semeraro2025}, \textit{Forma Mentis Network} \cite{stella2020text}, \textit{Ego Network} \cite{abramski2023cognitive}, and \textit{Mindset Stream} \cite{brian2023introducing} — and, where applicable, specifies a focal keyword for the ego network and two keywords to define the endpoints of the mindset stream. 

Upon clicking \textit{Draw Figures}, the dashboard constructs the TFMN by injecting the persona's edge list directly into a \texttt{FormamentisNetwork} object via the \texttt{\_replace()} method, and subsequently invokes the corresponding \texttt{emoatlas} rendering functions: \texttt{draw\_formamentis\_flower()} for the emotional flower, \texttt{draw\_formamentis()} for the full network and ego network (with the \texttt{highlight} parameter set to the focal keyword), \texttt{extract\_word\_from\_formamentis()} to isolate the ego subgraph, and \texttt{plot\_mindset\_stream()} to trace the conceptual path between the two selected keywords. Each figure is annotated with the full sociodemographic profile of the selected persona, enabling direct visual comparison across different subpopulations and debate configurations.

\subsubsection*{Limitations and Ethics}

\textbf{Limitations of persona-conditioned generation.} Persona prompting is a useful experimental tool, but it is not a transparent window into the attitudes of real human social groups \cite{de2025measuring,zheng2024helpful}. Generated outputs may reflect prompt wording, training data biases, safety tuning, and model-specific defaults \cite{zheng2024helpful,Hui2025}. For this reason, CDS should not be used as a substitute for human survey data, interview data, or behavioral measurements from human participants. The dataset is designed to be used for comparative analyses of model behavior and LLMs' bias detection under standardized contextual conditioning.

\textbf{Ethical use of the dataset.} CDS includes synthetic combinations of demographic, psychological, and social attributes across sensitive topics. Such combinations can amplify stereotypes or produce misleading generalizations if interpreted incautiously. We therefore recommend that users treat Cognitive Digital Shadows as a diagnostic resource for auditing LLM behavior, not as evidence about the beliefs of real populations. We also encourage downstream studies to report subgroup biases, reminiscent of human-based studies \cite{abramski2023cognitive}, examine sensitive outputs qualitatively \cite{DeDuro2025}, and avoid normative claims about human demographic groups based solely on CDS-generated text. CDSs should rather be used as shadows or counterparts to human behavior and could be integrated in social media platforms \cite{rossetti2024social,stella2020text}, educational contexts \cite{abramski2023cognitive} and more.

\section*{Data and Code availability}
The Python code and all related data used to produce the personas and validate the dataset are available at \url{https://github.com/NaviDATA-Repos/PENSO_Data_WP-ConvinceMe_FIS2_UniTrento.git}. The repository is organized into four main directories. 

\begin{figure}[h!]
\begin{forest}
  for tree={
    font=\ttfamily\small,
    grow'=0,
    child anchor=west,
    parent anchor=south,
    anchor=west,
    calign=first,
    edge path={
      \noexpand\path [draw, \forestoption{edge}]
      (!u.south west) +(7.5pt,0) |- node[fill,inner sep=1.25pt] {} (.child anchor)\forestoption{edge label};
    },
    before typesetting nodes={
      if n=1{insert before={[,phantom]}}{}
    },
    fit=band,
    before computing xy={l=15pt},
  }
[$PENSO\_Data\_WP-ConvinceMe\_FIS2\_UniTrento$/
  [Code/]
  [Data/
    [Processed\_Data/
      [Data\_visualization/ [{[19 LLM folders]}]]
      [EdgeList/ [{[19 LLM folders, some with PKL\_version\_of\_edgelists/]}]]
      [Hypothesis\_Testing/ [{[19 LLM folders]}]]
      [TFMN\_EmoA\_stats/]
    ]
    [Raw\_Data/ [{[19 LLM folders]}]]
  ]
  [Data Pooling System/]
]
\end{forest}
\end{figure}

\texttt{RawData\_LLM\_persona\_ConstraintsFixed/} contains the raw JSON persona records 
generated by each of the 19 LLMs, persona metadata, topic labels, model identifiers, opinion and reasoning-summary fields, organized per model. 
\texttt{ProcessedData\_ConstraintsFixes/} holds all analysis and validation outputs, subdivided into: 
\texttt{Data\_ visualization/} (figures and CSV files which are used for word count distributional analyses), 
\texttt{EdgeList/} (TFMN edge-list files per model), 
\texttt{Hypothesis\_Testing/} (KW statistical test outputs per model), and 
\texttt{TFMN\_EmoA\_stats/} (TFMN analysis, network features and aggregated statistics). 
\texttt{Code/} contains the Python scripts used for data processing, network construction, and figure generation for this article. 
\texttt{Data pooling dashboard/} contains the Google Colab dashboard without associated metadata (also archived at  \url{https://doi.org/10.5281/zenodo.19816544})

\subsection*{Online Data Pooling Dashboard}

The Google Colab pooling dashboard and associated metadata are accessible at \url{https://drive.google.com/drive/folders/16FJEFeFgiWe5CmY9q7--jonmfH39qfCt?usp=sharing}. This allows for the loading of the CDS metadata and the TFMN edge list data file that have been released, along with a dashboard for filtering the data based on topic, model, role mode, persona attributes, while producing visualisations of the TFMN, emotional flowers, ego network for selected words, and mindset streams as well. 

\section*{Acknowledgements}

This work was supported by the contribution of the Ministero dell'Università e della Ricerca (MUR) according to Decreto N. 23178 of 10 December 2024 - Bando FIS 2. The authors acknowledge support from CALCOLO, funded by Fondazione VRT, for support with the computational infrastructure simulating LLMs.

\bibliography{main}



\section*{Author contributions statement}

Both authors designed the dataset, conceptualized the experiments, and wrote the manuscript. M.S. generated the data, while A.A.A. curated the data, conducted the experiments, analyzed the results, and developed the dashboard. Both authors contributed to validation and manuscript review.


\section*{Competing interests} 
The authors declare no competing interests.

\end{document}